%% file: main.tex
\documentclass[10pt,twocolumn,letterpaper]{article}

\usepackage{iccv}
\usepackage{times}
\usepackage{epsfig}
\usepackage{graphicx}
\usepackage{amsmath}
\usepackage{amssymb}

\usepackage{bbm}

\usepackage{array,multirow}
\usepackage{mathtools}
\usepackage{xcolor}
\usepackage{dirtytalk}
\usepackage{comment}
\usepackage{algorithm}
\usepackage{listings}
\usepackage{nicefrac}
\usepackage{multirow}

\usepackage{etoolbox}
\makeatletter
\AfterEndEnvironment{algorithm}{\let\@algcomment\relax}
\AtEndEnvironment{algorithm}{\kern2pt\hrule\relax\vskip3pt\@algcomment}
\let\@algcomment\relax
\newcommand\algcomment[1]{\def\@algcomment{\footnotesize#1}}
\renewcommand\fs@ruled{\def\@fs@cfont{\bfseries}\let\@fs@capt\floatc@ruled
  \def\@fs@pre{\hrule height.8pt depth0pt \kern2pt}%
  \def\@fs@post{}%
  \def\@fs@mid{\kern2pt\hrule\kern2pt}%
  \let\@fs@iftopcapt\iftrue}
\makeatother

\newcommand{\juan}[1]{\textcolor{red}{{\textbf{Juan:} \emph{#1}}}}

\usepackage[pagebackref=true,breaklinks=true,letterpaper=true,colorlinks,bookmarks=false]{hyperref}

\iccvfinalcopy 

\def\iccvPaperID{1862} 

\ificcvfinal\fi

\begin{document}

\title{Enhancing Adversarial Robustness via Test-time Transformation Ensembling}

\author{
\normalsize Juan~C. Pérez$^{1,2}$
 , Motasem Alfarra$^{1}$
 , Guillaume Jeanneret$^{2}$
 , Laura Rueda$^{2}$,
 \\
\normalsize Ali Thabet$^{1}$, Bernard Ghanem$^{1}$, and Pablo Arbeláez$^{2}$ \\ 
\normalsize $^{1}$King Abdullah University of Science and Technology (KAUST),\\ 
\normalsize $^{2}$Center for Research and Formation in Artificial Intelligence, Universidad de los Andes,\\
\small $^{1}$\{\texttt{juan.perezsantamaria,motasem.alfarra,ali.thabet,bernard.ghanem}\}\texttt{@kaust.edu.sa}\\
\small $^{2}$\{\texttt{g.jeanneret10,l.ruedag,pa.arbelaez}\}\texttt{@uniandes.edu.co;}
}

\maketitle
\ificcvfinal\thispagestyle{empty}\fi

\input{Sections/abstract}
\vspace{-0.5cm}
\input{Sections/introduction}
\input{Sections/related_work}

\input{Sections/methodology}
\input{Sections/experiments}
\input{Sections/conclusions}

\textbf{Acknowledgments.} This work was partially supported by the King Abdullah University of Science and Technology (KAUST) Office of Sponsored Research.

{\small
\bibliographystyle{ieee_fullname}

\input{main.bbl}
}
\clearpage
\newpage
\input{Sections/suppl}

\end{document}

%% file: Sections/abstract.tex
\begin{abstract}
Deep learning models are prone to being fooled by imperceptible perturbations known as adversarial attacks. In this work, we study how equipping models with Test-time Transformation Ensembling (TTE) can work as a reliable defense against such attacks. While transforming the input data, both at train and test times, is known to enhance model performance, its effects on adversarial robustness have not been studied. Here, we present a comprehensive empirical study of the impact of TTE, in the form of widely-used image transforms, on adversarial robustness. We show that TTE consistently improves model robustness against a variety of powerful attacks without any need for re-training, and that this improvement comes at virtually no trade-off with accuracy on clean samples. Finally, we show that the benefits of TTE transfer even to the certified robustness domain, in which TTE provides sizable and consistent improvements.

\end{abstract}

%% file: Sections/introduction.tex
\section{Introduction}
The onset of deep learning techniques has revolutionized several fields such as Computer Vision~\cite{krizhevsky2012imagenet}, Natural Language Processing~\cite{mikolov2013efficient}, and Reinforcement Learning~\cite{mnih2013playing}. In the realm of computer vision, deep learning based methods have even surpassed
human-level performance on challenging datasets~\cite{he2015delving}. 
Despite the success of deep learning-based systems, researchers have noticed severe brittleness in their output: while remarkably accurate, they are extremely sensitive to imperceptible perturbations, now known as adversarial attacks~\cite{szegedy2014intriguing}.
\begin{figure}[t]
    \centerline{
    \includegraphics[width=\columnwidth]{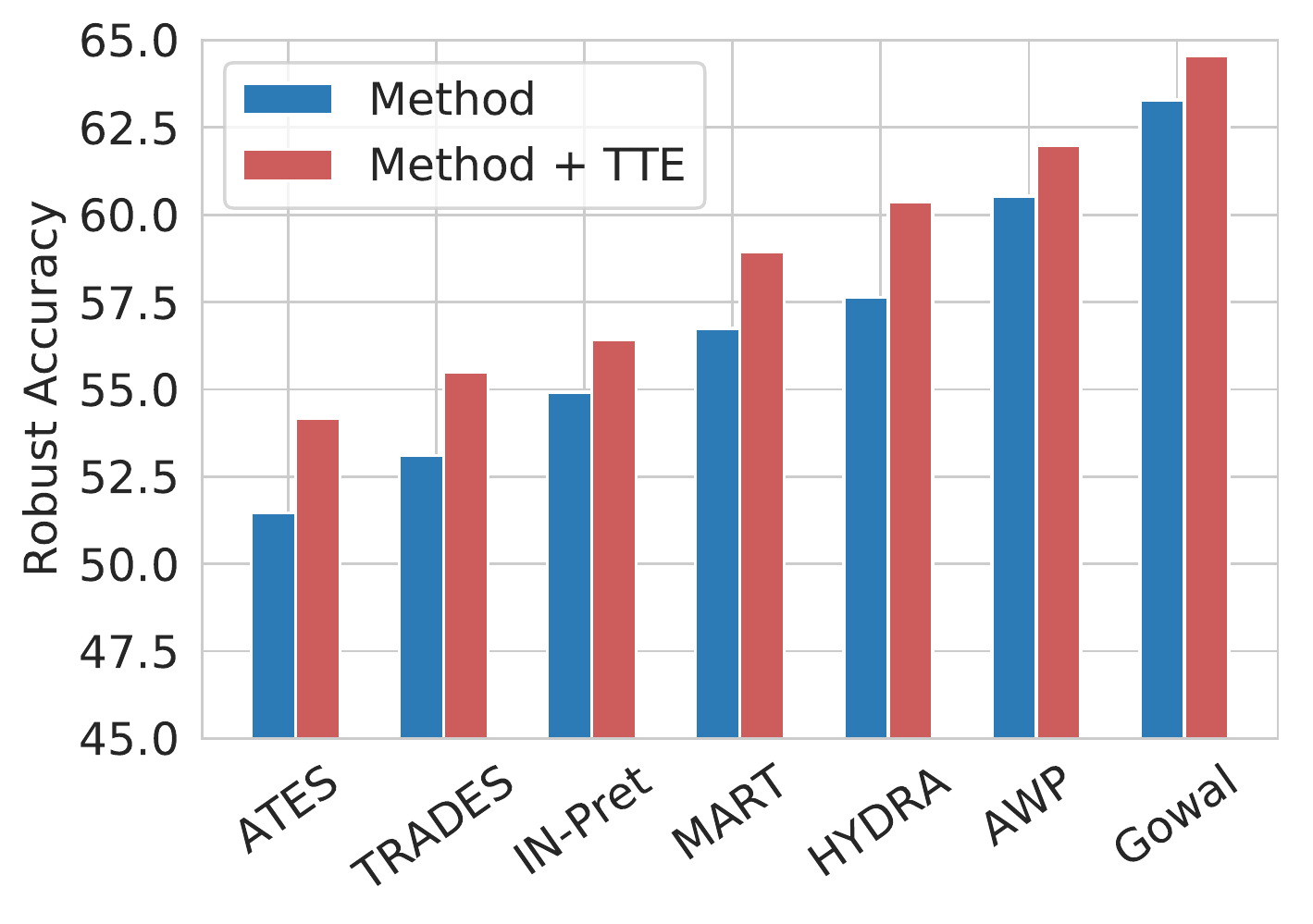}}
    \caption{\textbf{\textit{Test-time Transformation Ensembling} (TTE) enhances adversarial robustness.} Introducing TTE increases the adversarial robustness of several state-of-the-art defenses on CIFAR10.}
    \label{fig:pull}
\end{figure}

The discovery of this adversarial vulnerability, accordingly, has led to a large number of works addressing security concerns by proposing methods to defend models against attacks~\cite{carlini2017towards,goodfellow2015explaining,madry2018towards,gilmer2019adversarial}. Beyond such security concerns, adversarial vulnerability also sheds doubt on the nature of the impressive performance that computer vision systems achieve. While some of these systems may have been inspired by the human visual system, their lack of robustness could suggest that their inner workings strongly differ from those of humans~\cite{Xie2019CVPR}.

We highlight that current recognition methods perform predictions on single and static images. While this experimental setup is practical, it strongly differs from the way in which humans recognize objects in the real world. We note that this mismatch in how models perform predictions may be a factor contributing to the adversarial vulnerability:
studies in developmental psychology have noted how toddlers jointly exploit object permanence~\cite{piaget2008psychology, bremner1994infancy} and multiple views of objects to develop efficient, accurate and \textit{robust} visual systems \cite{slone2019self, mcquillan2020parents, jayaraman2019faces, perry2010learn, bambach2018toddler}. That is, while humans may perform recognition by ensembling multiple views of the intended object, machine learning-based systems traditionally focus on a single view. Inspired by these observations, we argue that ensembling predictions of transformed versions of an image can be cast as a simple and coarse simulation of how humans recognize objects in the real world.

Indeed, studies have shown that harnessing transform-based augmentation of data, both at train and test time, can provide consistent performance boosts in tasks such as detection~\cite{fcis_li17} and recognition~\cite{krizhevsky2012imagenet, vgg_simonyan15, he2016deep}. Other works have shown that randomization-based transforms and noise injection can be exploited to improve adversarial robustness~\cite{raff2019barrage, liu2018towards}, providing evidence that randomization is useful for defending against adversaries.

In this work, we study the effect that Test-time Transformation Ensembling (TTE) of model predictions on \textit{deterministically}-transformed versions of an image has on adversarial robustness. Specifically, we explore ensembling predictions over transformed versions of an image by applying two semantics-preserving transforms customarily used in computer vision (\textit{crops} and \textit{flips}). We present a comprehensive empirical study assessing the impact of introducing TTE against state-of-the-art adversaries. Our experiments show that equipping deep networks with TTE provides consistent improvements in adversarial robustness across datasets and attacks, while coming at virtually no cost to accuracy on clean samples. TTE is modular and versatile, and we show how it improves robustness of state-of-the-art defenses across-the-board (Figure~\ref{fig:pull}). We further validate the versatility of TTE by showing how it can also boost certified robustness. 

\textbf{Contributions:} \textbf{(1)} We present Test-time Transformation Ensembling (TTE), where we apply two semantics-preserving transforms (\textit{crops} and \textit{flips}), to enhance network robustness against adversaries. Upgrading a model with TTE does not need re-training and it can be implemented with less than $30$ lines of code. To ensure reproducibility, we will make our full implementation in PyTorch~\cite{paszke2019pytorch} publicly available \footnote{Code is available at \url{https://github.com/juancprzs/TTE}.}
\textbf{(2)} We show how TTE consistently provides significant improvements in adversarial robustness for top-performing methods on the \textit{AutoAttack} benchmark~\cite{croce2020reliable}, reaching improvements of  $2.04\%$ and $1.29\%$ on CIFAR10 and CIFAR100, respectively (Figure~\ref{fig:pull}). TTE's benefits also extend to the large-scale ImageNet dataset~\cite{deng2009imagenet}, where we enhance the state-of-the-art defense, Feature Denoising~\cite{Xie2019CVPR}, and conduct the first large-scale evaluation of adversarial robustness against AutoAttack on ImageNet. We find that, even in the ImageNet domain, TTE can provide boosts of $2.21\%$. 
\textbf{(3)} We find that the benefits of TTE are not confined to the \textit{empirical} assessment of adversarial robustness: TTE can also boost \textit{certified} defenses. Specifically, on CIFAR10, introducing TTE into Randomized Smoothing~\cite{cohen2019certified} and SmoothAdv~\cite{salman2019provably} provides a boost of \textbf{$10\%$} and \textbf{$7\%$} in Average Certified Radius, respectively.

%% file: Sections/related_work.tex
\section{Related work}
\textbf{Data transforms.} 
Classic transforms are behind most augmentation techniques for preventing overfitting of Deep Neural Networks~\cite{krizhevsky2012imagenet}. These transforms are now a \textit{de facto} standard in training-routine implementations, and are of such importance that automated learning of augmentation policies has been studied~\cite{cubuk2019autoaugment}. While transforms are effective during training, a stream of works also showed their benefits at test time~\cite{he2016deep,43022,vgg_simonyan15,krizhevsky2012imagenet,kim2020learningloss,Moshkov2020}. For instance, Sermanet \emph{et al.}~\cite{sermanet2014overfeat} augment the input at test time by including multi-scale information. Moreover,~\cite{fcis_li17,Zagoruyko2016Multipath} enhanced detectors by ensembling the predictions on both the original images and their transformed versions. In this work, we observe that data transforms are related to the \say{self-generated variability} phenomenon observed in infants~\cite{slone2019self}.
Inspired by this connection, we present a comprehensive empirical study of the effects of test-time transforms on adversarial robustness.

\textbf{Adversarial Robustness}. Robustness to adversarial noise is a pressing concern in the deep learning community~\cite{goodfellow2015explaining,dong2020benchmarking}. However, evaluating adversarial robustness has proven to be extremely difficult~\cite{carlini2019evaluating}. A large amount of works have proposed both white-box attacks~\cite{kurakin2017adversarial, madry2018towards, moosavi2016deepfool} and black-box attacks \cite{andriushchenko2020square, athalye2018obfuscated, athalye2018synthesizing, dong2019efficient} to evaluate adversarial robustness. Recently, Croce \textit{et al.}~\cite{croce2020reliable} proposed AutoAttack, an ensemble of four attacks with automatic hyper-parameter calibration that provides reliable assessments of adversarial robustness. From the side of defenses, a stream of works have aimed at developing models that are resistant to attacks. These works include adversarial training \cite{goodfellow2015explaining, madry2018towards}, combining adversarial training with pre-training on external data sources \cite{hendrycks2019using, chen2020adversarial, kim2020adversarial, carmon2019unlabeled}, and TRADES \cite{zhang2019theoretically}, among many others \cite{wu2020adversarial, wang2019improving, sehwag2020hydraPruning, chen2020efficient}. In this work, we study how defenses can be equipped with TTE to boost adversarial robustness as measured via AutoAttack.

\textbf{Certified Defenses.} The aforementioned empirical assessment of adversarial robustness cannot provide guarantees about the inexistence of adversarial examples for a given classifier.
This concern has incited interest in certified adversarial robustness, which aims at developing models whose predictions can be certified, \textit{i.e.} predictions that are verifiably constant within some radius around any input~\cite{wong2018provable, raghunathan2018certified}. Here, we note that many approaches have been proposed for the purpose of certification ~\cite{zhai2020macer, cohen2019certified, salman2019provably}. 
In this paper, we study the impact of TTE on certification by experimenting with two well-known certified defenses: Randomized Smoothing~\cite{cohen2019certified} and SmoothAdv~\cite{salman2019provably}. Our experiments find that TTE's benefits also appear in certification, where TTE can boost these defenses by sizable margins.

%% file: Sections/methodology.tex
\section{Methodology}\label{sec:methodology}

\input{algorithms/TTE_wrapper}
We study the impact of introducing TTE on adversarial robustness. We observe that traditional image transforms can provide simple transforms to be used in TTE. Thus, we aim at studying the impact that introducing customary image transforms at test time has on adversarial robustness. In particular, we set to study a simple TTE wrapper for trained classification models. This wrapper is a module that receives an image, augments the image with a fixed set of transforms, inputs the image to the model, and, finally, ensembles the outputs through averaging. 

Our selection of transforms for the wrapper is based on facilitating the assessment of adversarial robustness of the wrapped model. Specifically, we are interested in wrapping models and then properly conducting white-box attacks, since these attacks are at the core of the most challenging adversarial settings. The transforms we select, thus, must not hinder nor obstruct the computation of the gradient with respect to the model's input. Hence, we select three well-known label-preserving transforms that are common in training routines: \textit{(i)} horizontal flips, \textit{(ii)} padding-and-cropping, and \textit{(iii)} the composition of these two.

Note that these transforms are easily implemented in a \textbf{differentiable and deterministic} manner: all transforms are implemented by indexing the input tensor, which, during the backward pass, translates to simply directing the gradient to the selected indices. Further, we remark that we refrain from introducing stochasticity in the transforms, contrary to common implementations of training and testing routines. That is, the set of transforms and their parameters are kept \textit{fixed} after initialization: flips are performed deterministically, and the sections being cropped are permanent. Once the set of transforms and their parameters have been set, we instantiate the transforms and use them for initializing the wrapper. Hence, we emphasize that our approach does not hinder nor obstruct computing the gradient of the wrapper's output with respect to its input. Please refer to Algorithm~\ref{alg:wrapper} for pseudo-code of the wrapper, and to the \textbf{Supplementary Material} for the pseudo-codes of the transforms we study.

In the following section, we conduct a comprehensive empirical study of the adversarial robustness effects of equipping models with TTE.
Remarkably, we find that TTE can consistently improve adversarial robustness across defenses, while requiring less than $30$ lines of code. 

%% file: algorithms/TTE_wrapper.tex
\begin{algorithm}[t]
\caption{TTE Wrapper pseudo-code in PyTorch style.}
\label{alg:wrapper}
\algcomment{\fontsize{7.2pt}{0em}\selectfont \texttt{cat}: concatenation.
}
\definecolor{codeblue}{rgb}{0.25,0.5,0.5}
\lstset{
  backgroundcolor=\color{white},
  basicstyle=\fontsize{7.2pt}{7.2pt}\ttfamily\selectfont,
  columns=fullflexible,
  breaklines=true,
  captionpos=b,
  commentstyle=\fontsize{7.2pt}{7.2pt}\color{codeblue},
  keywordstyle=\fontsize{7.2pt}{7.2pt},
}
\begin{lstlisting}[language=python]
# C: number of channels
# H: image height
# W: image width
# N: number of classes
class TTEWrapper:
    def init(self, model, transforms):
        # transforms: list of differentiable functions
        self.model = model
        self.transforms = [lambda x: x] # the identity
        self.transforms += transforms
    
    def forward(self, x):
        # apply transforms
        x = cat([t(x) for t in self.transforms])
        # move transforms to the batch dimension
        x = x.view(-1, C, H, W)
        # forward
        s = self.model(x)
        # move scores of transforms to other dimension
        s = s.view(len(self.transforms), -1, N)
        # average scores across transforms dimension
        return s.mean(dim=0)
\end{lstlisting}
\end{algorithm}

%% file: Sections/experiments.tex
\section{Experiments}
We conduct a comprehensive empirical study regarding the effects of TTE on adversarial robustness. We vary defenses, datasets, and attacks, while also gathering insights into TTE's inner workings. We find that TTE yields sizable and consistent boosts in robustness, both against strong attacks and for certification purposes.

\subsection{Adversarial Robustness Assessment}
A reliable assessment of adversarial robustness is fundamental to our study. Hence, we use the challenging AutoAttack benchmark~\cite{croce2020reliable} for estimating adversarial robustness. AutoAttack is a parameter-free ensemble of diverse attacks that has shown outstanding capabilities of identifying vulnerabilities in adversarial robustness defenses. AutoAttack is composed of: AutoPGD, aiming at optimizing either Cross Entropy (CE) or a targeted version of a Difference-of-Logits-Ratio~\cite{croce2020reliable} (denoted as \textit{APGD-CE} and \textit{APGD-T}, respectively); targeted Fast Adaptive Boundary~\cite{croce2020minimally} (denoted as \textit{FAB-T}) that aims at perturbation minimization and has shown promising results against gradient-masking; and Square Attack~\cite{andriushchenko2020square}, a norm-bounded score-based black-box attack that does not rely on gradient information (denoted as \textit{Square}). High performance against such a diverse and powerful ensemble has shown to provide an accurate assessment of adversarial robustness~\cite{croce2020reliable}. 

For all experiments, we run AutoAttack and report both clean accuracy and \textit{Robust} accuracy, where the latter is defined as the per-instance worst case across all attacks. For the main experiments and with completeness in mind, we also report the accuracy against individual attacks composing AutoAttack. 

\input{tables/normally_trained_model}

\subsection{TTE on Undefended Models}
We begin our study by analyzing the effect of introducing TTE on the adversarial robustness of undefended models. We study this setting by conducting nominal training of a ResNet-18~\cite{he2016deep} on CIFAR10 and CIFAR100~\cite{krizhevsky2009learning}, and then attacking these models both with and without TTE. The attack strength values ($\epsilon$) that the robustness community usually employs are capable of dropping the accuracy of undefended models to approximately $0$\%~\cite{carlini2017towards}.
Thus, for this experiment, we use a weaker attack strength of $\epsilon=\nicefrac{2}{255}$ that allows us to observe variations in adversarial robustness. Table~\ref{tab:undefended} summarizes the results. For both datasets, we find consistent increments in both clean and robust accuracy. In particular, robust accuracy increases by a remarkable $13\%$ on CIFAR10, and by $\approx 1\%$ on CIFAR100. Our results thus provide evidence that TTE is capable of boosting the adversarial robustness of models without hurting accuracy on clean samples. 

\subsection{Boosting the State-of-the-Art}\label{subsec:sota}
\input{tables/SOA_cifar10_and_100}
\input{tables/ImageNet}

\textbf{CIFAR.} The AutoAttack benchmark hosts an online leaderboard\footnote{Available at \url{https://github.com/fra31/auto-attack}}, where current state-of-the-art (SOTA) defenses on various datasets are ranked according to their performance against AutoAttack. To evaluate the impact of equipping SOTA defenses with TTE, we download pre-trained models of defense approaches from the CIFAR10 leaderboard, add our TTE wrapper, and run AutoAttack on them with a perturbation budget of $\epsilon=\nicefrac{8}{255}$. Our selection of defenses for evaluation is based on high-performance in the AutoAttack benchmark and availability of the trained models (either from the official repository or upon contacting the respective authors). In particular, we test TTE on six  high-performing defenses: Adversarial Training with Early Stopping (ATES)~\cite{sitawarin2020improving}, Tradeoff-inspired Adversarial Defense via Surrogate-loss minimization (TRADES)~\cite{zhang2019theoretically}, ImageNet pre-training (IN-Pret)~\cite{hendrycks2019using}, robust-network pruning (HYDRA)~\cite{sehwag2020hydraPruning}, Misclassification Aware Adversarial Training (MART)~\cite{wang2019improving}, Adversarial Weight Perturbation (AWP)~\cite{wu2020adversarial}, and the method of Gowal \textit{et al.}~\cite{gowal2020uncovering}. Note that only three of these defenses are available for CIFAR100, as this dataset is much less studied for adversarial robustness. 

We report the performance of the standard and TTE-enhanced versions of these defenses in Table~\ref{tab:cifar10}. In line with previous works~\cite{krizhevsky2012imagenet, vgg_simonyan15, he2016deep}, our results show how equipping models with TTE generally enhances their clean accuracy. More importantly, we observe that enhancing defenses with TTE consistently increases the robustness of \textit{all} defenses across both attacks and datasets. In CIFAR10, for instance, TTE increases robust accuracy by $2.04\%$ on average. Analogously, in CIFAR100, we observe an average increase of $1.29\%$.

Note that the strongest defense for CIFAR10 is Gowal \textit{et al.}, and for CIFAR100 the strongest defense is AWP. Equipping these top-performing defenses with TTE increases their robust accuracy by $1.26\%$ and $0.86\%$ on CIFAR10 and CIFAR100, respectively.

\textbf{ImageNet.} Adversarial robustness defenses in ImageNet~\cite{deng2009imagenet} are much less common than in CIFAR, mainly due to the computational costs associated with conducting adversarial attacks on such large images and vast amount of instances. A recent robustness assessment by Dong \textit{et al.}~\cite{dong2020benchmarking} suggests that the SOTA defense on ImageNet is Feature Denoising (FD)~\cite{Xie2019CVPR}, which combines large-scale adversarial training~\cite{madry2018towards} with the introduction of denoising blocks into the model's architecture. We download the pre-trained models from the official implementation and run AutoAttack on the standard and TTE-enhanced versions\footnote{Since this model was not trained on zero-padded images, we extract $224\times224$ crops from the $256\times256$ center crop of the resized image. We resize the shortest side to $256$ while preserving the height-width ratio.}. Following the methodology of~\cite{Xie2019CVPR}, we run attacks with a perturbation budget of $\epsilon = \nicefrac{16}{255}$. 

We report the results of this experiment in Table~\ref{tab:imagenet}. These results show that, even in the large-scale regime of ImageNet, TTE is still able to provide improvements in adversarial robustness. In fact, we observe consistent increase in accuracy against every attack of the AutoAttack benchmark. In particular, we find that TTE is able to provide a remarkable improvement of $2.21\%$ in robust accuracy. It is worthwhile to mention that we report the results corresponding to the best TTE selection, while leaving the extensive ablation results to the \textbf{supplementary material}.

\subsection{Transform Selection}\label{subsec:transformation}
In Section~\ref{sec:methodology}, we established that the transforms considered in this study are horizontal flips, pad-and-crop, and the composition of these two. These transforms were selected so as to facilitate the assessment of adversarial robustness by introducing transforms that are both differentiable and deterministic. Since the padding-and-cropping operation is parameterized by the indices of the crop (the padding size is another parameter, but in our study we kept it fixed to $4$), there is a large number of transforms that could be generated from an image. The combination of such transforms with horizontal flipping yields a vast search space. In this section, we study a simple set of these transforms, and show that a large majority of transforms already provides sizable gains in adversarial robustness.

As subjects of our study, we choose TRADES~\cite{zhang2019theoretically} for experiments on CIFAR10 and FD~\cite{Xie2019CVPR} for experiments on ImageNet. We use the wrapper introduced in Section~\ref{sec:methodology} around both models (as obtained from their official implementations), and test various selections of transforms for initializing the wrapper. In particular, we consider: \textit{(i)} flip, \textit{(ii)} crops, \textit{(iii)} flip + crops, and \textit{(iv)} flip + crops + flipped-crops. Whenever we use a crop, we vary the number of extracted crops from one to four. For all the cases that include crops, we randomly \textit{initialize} the indices of the crop, and keep them fixed after, \textit{i.e.} note that the transform is, again, \textit{deterministic}.

We report the evaluation of these transforms for TRADES in Table~\ref{tab:transformations} and for FD\footnote{Due to computational costs, we test FD with the official implementation's 30-step PDG attack (PGD$^{30}$) under the $\ell_\infty$ norm with $\epsilon=\nicefrac{16}{255}$.} in Table~\ref{tab:imagenet-transformations}. Our results show that \textit{any} of the studied transforms provides gains in adversarial robustness. Further, we note that even simple transforms already provide sizable gains. For instance, simply ensembling the original input with its flipped version boosts robust accuracies by $1.70\%$ and $0.97\%$ for TRADES and FD, respectively. For TRADES, we obtain the largest increase both in clean and robust accuracies with an ensemble composed of a flip, three crops, and three flipped-crops. This particular set of transforms achieves a remarkable boost of $2.38\%$ in robust accuracy. However, we report the largest gain in robust accuracy ($+1.98\%$) for FD when the image is processed jointly with four crops, four flipped-crops and the flipped instance. Our results show that TTE provides consistent gains in adversarial robustness across a simple set of transforms for both benchmarks. This outcome suggests that, despite the large size of the space of possible transforms, finding a set of transforms that provides sizable gains in adversarial robustness is effortless. In particular, in our experiments, \textit{all} the transforms provided gains in adversarial robustness.

\input{tables/ablation_on_several_transformations}

\input{tables/ImageNet-transforms}

\subsection{Are all crops created equal?} \label{subsec:equalcrops}
In Section~\ref{subsec:transformation}, we studied how several transforms affect adversarial robustness. From Tables~\ref{tab:transformations} and \ref{tab:imagenet-transformations}, we notice an odd phenomenon: increasing the number of crops does not always yield larger robustness gains. Since the crop location is random, this phenomenon suggests that there are crops that provide larger robustness gains than others.

We conduct an exhaustive search over the possible crops, and record changes in clean and robust accuracy. As mentioned in Section~\ref{subsec:transformation}, we fix the padding size to $4$ in the pad-and-crop transform. Hence, the height and width of each image increases by $8$ (each side increases by $4$). Thus, the total number of possible crops is $(8 + 1)^2 = 81$. 

We conduct this experiment on TRADES and evaluate on CIFAR10. Since the evaluation of each model is computationally expensive, we refrain from using the official TRADES model (a large WideResNet-34-10~\cite{zagoruyko2016wide} model), and rather train a smaller ResNet-18~\cite{he2016deep} model with the official TRADES implementation. This trained model achieves a clean accuracy of $80.96\%$ and a robust accuracy of $48.64\%$ against AutoAttack. For studying the impact of specific crops on robustness, we are interested in the evaluation of the model under two settings: \textit{(i)} when the model's input is \textit{only} the selected crop, \textit{i.e.} the model is not given the original image, and \textit{(ii)} when the model's input is both the original image \textit{and} the crop, that is, TTE itself.

\begin{figure*}[ht]
    \centering
    \includegraphics[width=1.0 \textwidth]{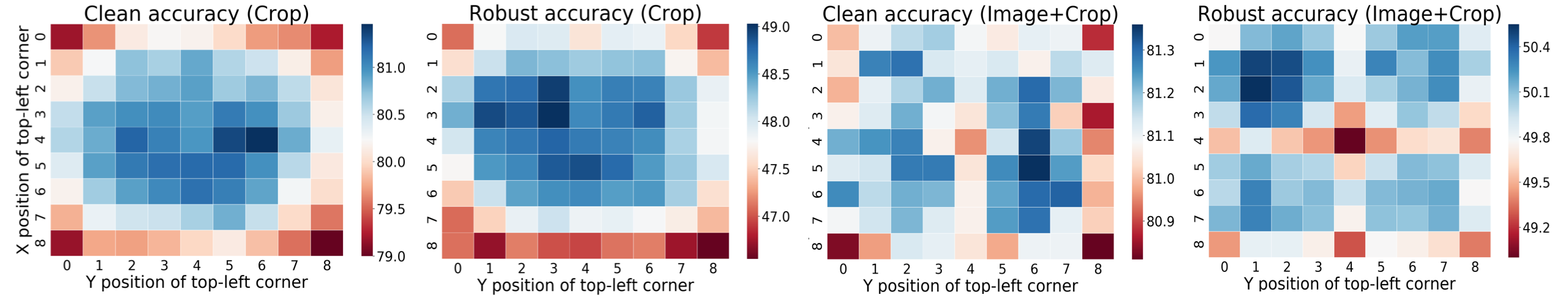}
    \caption{\textbf{Impact of specific crops on performance.} The heatmaps depict the clean and robust accuracies as a function of the location of the crops. The first two plots are obtained by only feeding cropped versions of CIFAR10 images, and the last two by feeding the original image \textit{and} its cropped version. We assessed a total of $81$ crops corresponding to all the possible translations in each scenario.}
    \label{fig:crops}
\end{figure*}

Figure~\ref{fig:crops} displays our results as heatmaps. The \verb'x' and \verb'y' coordinates of the heatmap range from $0$ through $8$, and are interpreted as the offset from the top-left corner of the zero-padded version of the image. These heatmaps show appealing spatial patterns. We notice that when using only the cropped versions of the images, spatial translations closer to the origin provide the best performance in both clean and robust accuracy. In particular, cropped images can even achieve better clean and robust accuracy than their uncropped versions under most one-pixel shifts. However, when using the original image \textit{and} its cropped version, the optimal crops shift towards the edges. In general, crops that arise from translations in both directions tend to consistently boost performance. Although a clear pattern in clean accuracy is elusive, robust accuracy tends to symmetrically improve as the selected crop moves towards the corners of the zero-padded image, with the exception of the corners themselves. This result implies that the model benefits from seeing the most shifted views of the image up to a certain threshold, but further shifts may be counterproductive.

\subsection{Is TTE obfuscating the gradient?}\label{sec:obf}
\input{tables/obfuscation_ablations}
In this paper, we construct the set of transforms to be studied based on a single criterion: that such transforms would ease the assessment of adversarial robustness. We consider transforms that would not hinder nor obstruct the computation of the gradient with respect to the input image. Thus, we study transforms that are differentiable and deterministic. Based on this criterion, we expect that adding the TTE wrapper does not lead to gradient obfuscation, which could yield an inaccurate assessment of adversarial robustness and, hence, a false sense of security~\cite{athalye2018obfuscated}. Here, we empirically test this hypothesis. In particular, we are interested in knowing whether TTE prevents the model from producing \say{useful gradients} for iterative optimization attacks \cite{papernot2017practical, athalye2018obfuscated}. Following~\cite{athalye2018obfuscated}, we check how the behavior of the model's performance changes when the iterative attack's parameters vary in terms of: \textit{(i)} number of optimization steps and \textit{(ii)} attack strength. 

This experiment requires a large number of runs. Hence, we follow the same experimental setup as in Section~\ref{subsec:equalcrops} and use a ResNet-18 model trained with TRADES. We compare this defense against its TTE-enhanced version that could cause, in theory, the largest effect on gradient obfuscation: original image + flip + 4 crops + 4 flipped-crops. Hence, we expect this experiment to provide an upper-bound to the (possible) gradient obfuscation effect of introducing TTE. We compute the adversarial accuracy of these defenses against the \textit{APGD-T}~\cite{croce2020reliable} attack from the AutoAttack ensemble, which is a strong iterative gradient-based adversary. We vary the number of optimization steps from $1$ to $100$ (the default number of steps for \textit{APGD-T} in AutoAttack), and the attack strength from $\nicefrac{1}{255}$ to $\nicefrac{64}{255}$. 

We report our main results in Table~\ref{tab:grad_obf}, and detail the rest in the \textbf{supplementary material}. We find that TTE suffers both when increasing the attack's number of iterations and its strength. Notably, the performance eventually reaches approximately $0\%$ when $\epsilon = \nicefrac{64}{255}$. Combined with the observation that white-box attacks are more successful than black-box attacks in fooling an TTE-enhanced model (as shown in Tables~\ref{tab:cifar10} and \ref{tab:imagenet}), these results suggest that introducing TTE does \textit{not} induce gradient obfuscation.

\begin{figure*}[t]
    \centerline{
        \includegraphics[width = 0.25\linewidth]{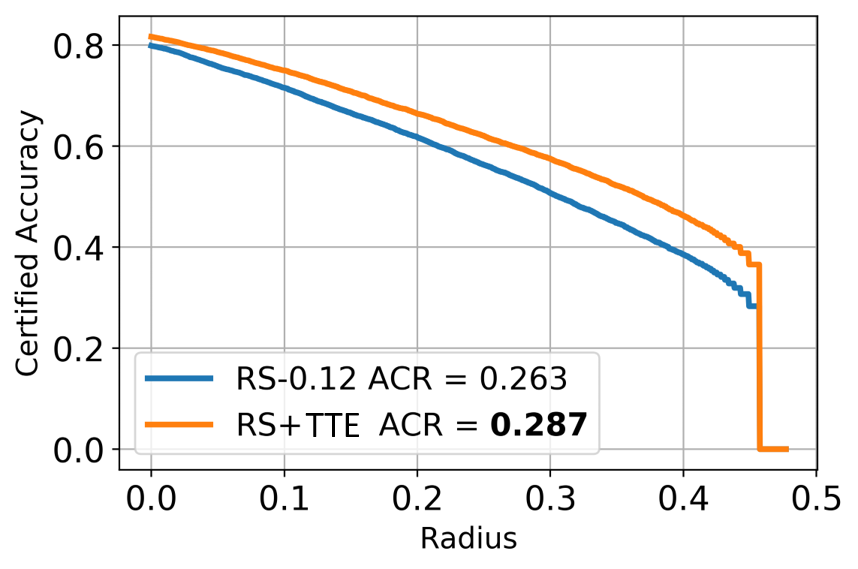}
        \includegraphics[width = 0.25\linewidth]{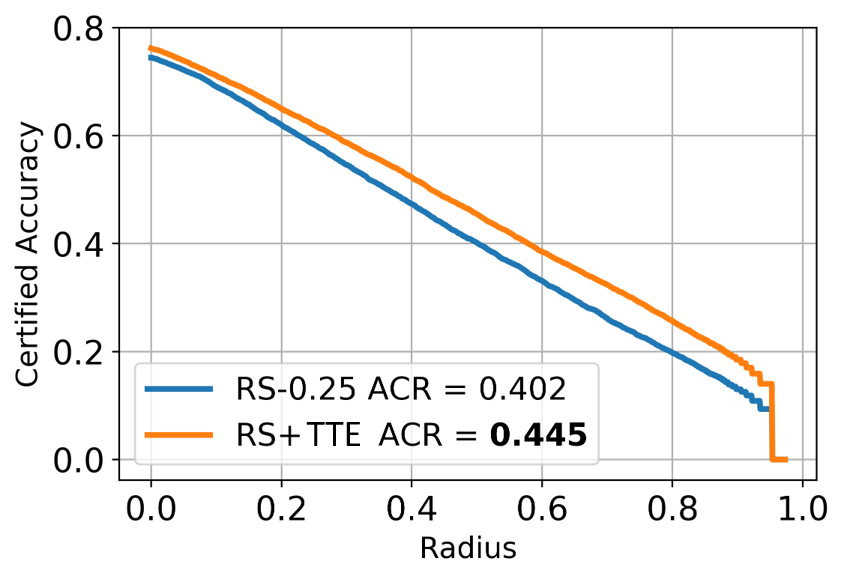}
        \includegraphics[width = 0.25\linewidth]{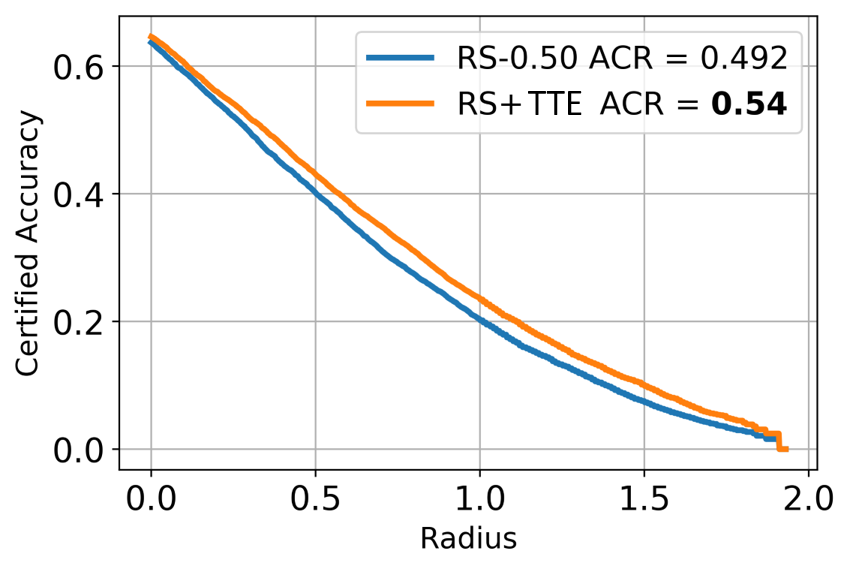}
        \includegraphics[width = 0.25\linewidth]{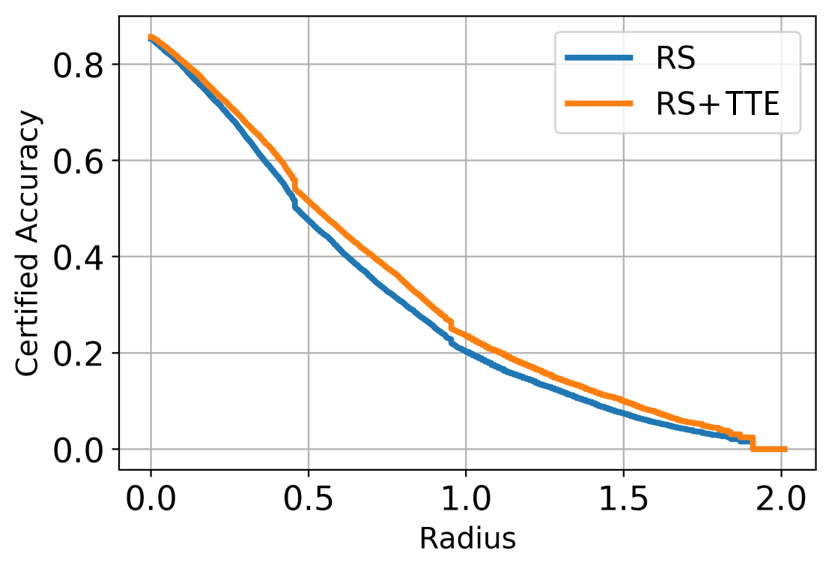}
    }
    \centerline{
        \includegraphics[width = 0.25\linewidth]{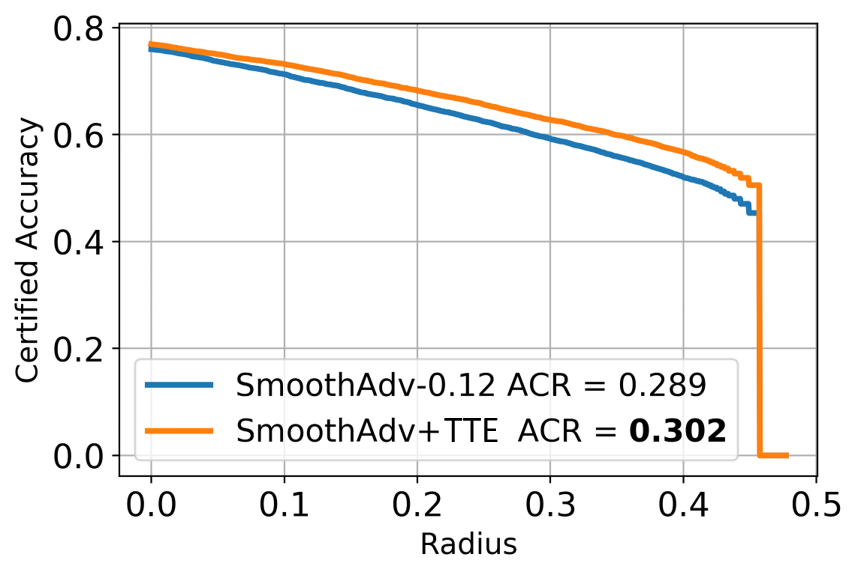}
        \includegraphics[width = 0.25\linewidth]{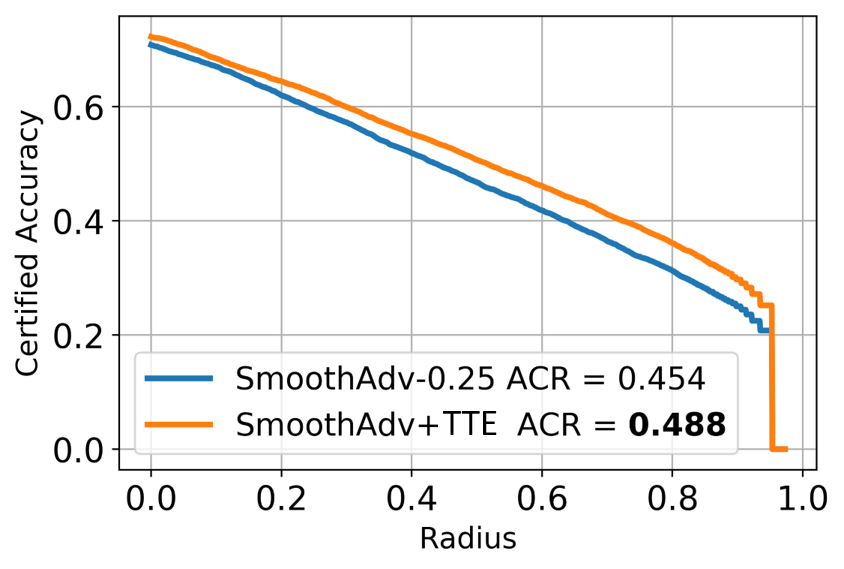}
        \includegraphics[width = 0.25\linewidth]{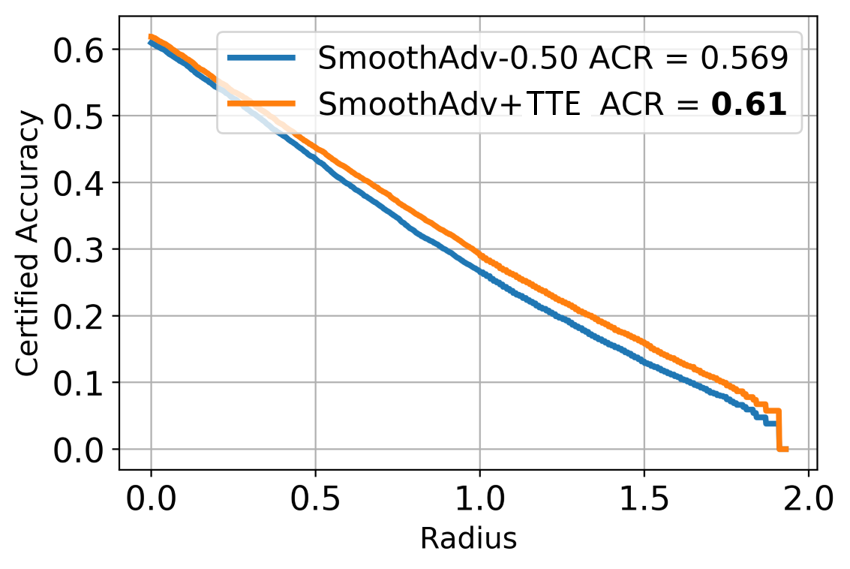}
        \includegraphics[width = 0.25\linewidth]{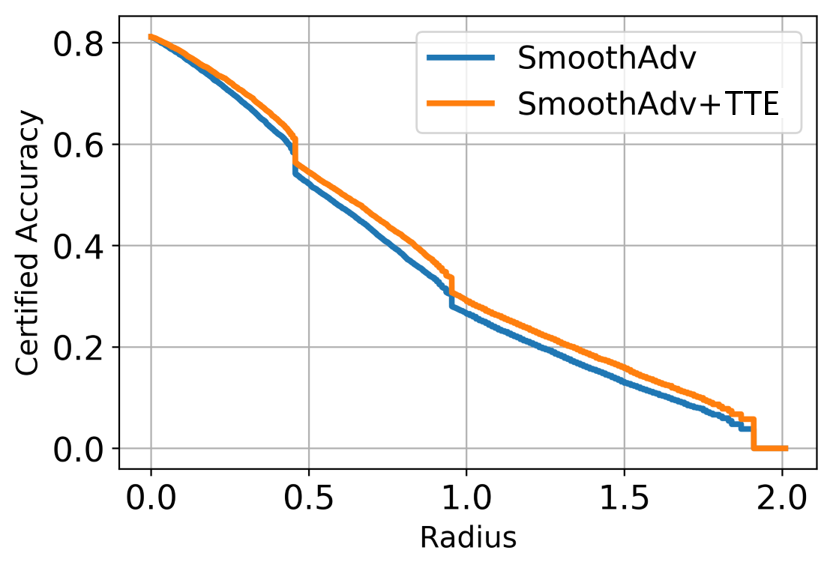}
    }
    \caption{\textbf{Boosting certified defenses with TTE.} We introduce TTE into two certified defenses: Randomized Smoothing (RS)~\cite{cohen2019certified} and SmoothAdv~\cite{salman2019provably} (top and bottom row, respectively). Each plot reports a certified accuracy curve. The first three columns show certified accuracy for $\sigma \in \{0.12, 0.25, 0.5\}$, respectively, and each legend reports the Average Certified Radius (ACR). The fourth column shows the envelope of the corresponding first three plots: the largest certificate that can be granted to each method across $\sigma$ values. }
    \label{fig:certified}
\end{figure*}

\subsection{Matching train- and test-time transforms}\label{sec:match}
Our experiments have considered three transforms: flip, pad-and-crop, and the combination of these two. We have shown how applying these transforms at \textit{test} time boosts the robustness of several defenses. We note that \textit{all} the defenses we studied also use these transforms at \textit{train} time, as is common practice in the training routines for image classifiers. Thus, in all our experiments, a match has existed between transforms on which defenses were trained and the ones we introduced at test time. Here, we experiment with introducing a mismatch between the train- and test-time transforms. Specifically, we study two setups: \textit{(i)} introducing a transform that is not seen during training (Gaussian filtering), and \textit{(ii)} training the model by \textit{removing} the pad-and-crop transform and then testing on padded-and-cropped images.

\textbf{Gaussian filtering.} We study the impact of introducing a Gaussian-filtering transform into TTE. We choose this transform based on spectral properties of adversarial examples, as recent works find a relation between high-frequency components and adversarial examples~\cite{wang2020high, yin2019fourier}. These findings suggest that using Gaussian filtering, a low-pass filter, could increase the adversarial robustness of a defense. Note that this transform can also be implemented in a differentiable manner, complying with our requirements. 

Thus, we modify TTE to only include Gaussian filtering, equip TRADES with TTE, and conduct attacks. We consider Gaussian filters with filter sizes $k \in \{3, 5\}$, and standard deviations $\sigma \in \{1, 2\}$. We report the results in Table~\ref{tab:nocrop_no_gauss} (top sub-table). Results indicate that introducing Gaussian filtering at test time, when the defense was \textit{not} trained on such transforms, is \textit{detrimental} to adversarial robustness.

\input{tables/without_pad_and_crop}

\textbf{Removing pad-and-crop from training.} We train a TRADES model by removing the usual pad-and-crop transform, and refer to this model as TRADES$^\text{nc}$. We record the performance of TRADES$^\text{nc}$ without TTE and with TTE based only on pad-and-crop transforms. We vary the number of crops from one to four. We report results in Table~\ref{tab:nocrop_no_gauss} (bottom sub-table). We note that \textit{(i)} the pad-and-crop train-time transform is important for TRADES: removing this transform decreases clean and robust accuracies, approximately, by $2\%$; and \textit{(ii)} adding a padded-and-cropped version of the image is detrimental to the adversarial robustness of TRADES$^\text{nc}$: \textit{any} number of additional transformed versions of the image hurt both clean and robust accuracies. Our results suggest that, for a TRADES model, there is large distribution mismatch between the original images and their padded-and-cropped versions.

The results in Table~\ref{tab:nocrop_no_gauss} indicate that matching train- and test-time transforms is fundamental for adversarial robustness. That is, simulating TTE through image transforms is beneficial for adversarial robustness, if and only if, a match exists between train- and test-time transforms.

\subsection{Boosting Certified Defenses}
Our experiments show that TTE provides improvements in the empirical assessment of adversarial robustness. Here, we test whether these improvements also transfer to the domain of certifiably robust models. Formally, for an input $x$ with true label $y$, a model $f$ is certifiably robust at $x$ with radius $R$ if $f(x) = f(x+\delta) = y \: ~~\forall \: \|\delta\|_p \leq R$. Thus, the certified accuracy of a model $f$ at a radius $R$ is defined as the portion of the test set for which the model is certifiably robust with a radius of at least $R$. Recently, two closely-related methods have been proposed for certification. Cohen \textit{et al.}~\cite{cohen2019certified} proposed Randomized Smoothing (RS), a technique that provides tight certified classification to an input by assigning the most probable class predicted by $f$ when $x$ is exposed to Gaussian noise of the form $\mathcal{N}(0, \sigma^2 I)$. Moreover, Salman \textit{et al.}~\cite{salman2019provably} proposed SmoothAdv, extending RS with adversarial training to boost certification. 

We study the effect of TTE on these defenses on CIFAR10. We follow~\cite{cohen2019certified} for RS and train a ResNet-18 with Gaussian augmentation. For SmoothAdv, we follow~\cite{salman2019provably} and apply Gaussian augmentation on PGD-generated images. We study certification in the $\ell_2$ sense and test with $\sigma \in \{0.12, 0.25, 0.5\}$. We introduce TTE to the models, certify the test set ($\mathcal S_{\text{test}}$), and report certification curves as well as the Average Certified Radius (ACR)~\cite{zhai2020macer}, defined as $\text{ACR} = \nicefrac{1}{|\mathcal S_{\text{test}}|} \sum _{(x,y) \in \mathcal S_{\text{test}}} R(x) \: \mathbbm{1}_{\{f(x) = y\}}$, where $\mathbbm{1}$ is the indicator function.

We summarize our results in Figure~\ref{fig:certified}. We report a curve per $\sigma$ and the curves' envelope, \ie the largest certificate that can be granted to each defense across all $\sigma$ values. We observe that TTE consistently improves certified accuracy for all radii, for all values of $\sigma$, and for both defenses. In particular, at $\sigma = 0.5$, TTE improves the ACR  by approximately $10\%$ and $7\%$ on RS and SmoothAdv, respectively. We remark that equipping these certified defenses with TTE comes at no cost during training, as TTE is only used at test time. That is, similar to our experiments on the empirical assessment of adversarial robustness, TTE provides boosts in the certification domain without requiring re-training.

Our results demonstrate that the benefits of TTE are not confined to the empirical assessment of robustness: TTE can also enhance \textit{certified} adversarial robustness. 

%% file: tables/normally_trained_model.tex
\begin{table}[]
\centering
\caption{\textbf{TTE and undefended models.} We test the impact of TTE on undefended models. TTE increases the adversarial robustness of undefended models by sizable margins. Best results in \textbf{bold}.
}
\centering
\small
\begin{tabular}{l|c|c|c}
\multicolumn{4}{c}{\textbf{CIFAR10}} \\ \hline
Method              & Clean         & Robust      & Diff.                   \\ \hline\hline
ResNet-18          & 92.58	        & 16.18       & \multirow{2}{*}{+13.63} \\ \cline{1-3}
ResNet-18 + TTE    & \bf{93.42}    & \bf{29.81}  &                         \\ \hline
\multicolumn{4}{c}{\textbf{CIFAR100}} \\ \hline
ResNet-18          & 76.66	        & 0.72       & \multirow{2}{*}{+0.96}  \\ \cline{1-3}
ResNet-18 + TTE    & \bf{77.57}    & \bf{1.68}  &                         \\ \hline
\end{tabular}
\label{tab:undefended}
\end{table}

%% file: tables/SOA_cifar10_and_100.tex
\begin{table*}[]
\centering
\small
\caption{\textbf{Adversarial robustness on CIFAR.} We compare the standard and the best TTE-enhanced versions of several defenses from the AutoAttack leaderboard, whose models are publicly available, on the CIFAR10 and CIFAR100 datasets. We report the clean accuracy, the adversarial accuracy against the individual attacks that compose AutoAttack, the worst-case (\textit{Robust}) accuracy in $(\%)$, and the average difference across datasets. Best results in clean and robust accuracies are shown in \textbf{boldface}. Note that \textit{all} defenses gain adversarial robustness when equipped with TTE.}
\centering
\begin{tabular}{c|l|c|cccc|c|c|c}
\hline
& Method                                      & Clean     & APGD-CE & APGD-T & FAB-T & Square & Robust    & Difference & Average              \\ \hline\hline
\parbox[t]{2mm}{\multirow{14}{*}{\rotatebox[origin=c]{90}{CIFAR10}}} & ATES~\cite{sitawarin2020improving}          & 86.84	    & 53.5	  & 51.5   & 51.91 & 59.77  & 51.46     & \multirow{2}{*}{+2.71}  & \multirow{14}{*}{+2.04} \\
& ATES + TTE                                  & \bf{86.86}& 56.48   &	54.19  & 54.70 & 60.67  & \bf{54.17}&                          \\ \cline{2-9}
& TRADES~\cite{zhang2019theoretically}        & 84.92     & 55.31   & 53.12  & 53.55 & 59.41  & 53.11     & \multirow{2}{*}{+2.38}   \\
& TRADES + TTE                                & \bf{85.14}& 57.46	  & 55.51  & 55.88 & 60.22  & \bf{55.49}&                          \\ \cline{2-9}
& IN-Pret~\cite{hendrycks2019using}   & 87.11     & 57.65   & 55.32  & 55.68 & 62.40  & 54.92     & \multirow{2}{*}{+1.51}   \\
& IN-Pret + TTE                       & \bf{87.13}& 59.06   & 56.44  & 56.73 & 63.14	& \bf{56.43}&                          \\ \cline{2-9}
& MART~\cite{wang2019improving}               & 87.50     & 62.18   & 56.80  & 57.34 & 64.87  & 56.75     & \multirow{2}{*}{+2.19}   \\
& MART + TTE                            & \bf{87.79}
& 63.95  & 58.94 & 59.51 & 65.62 & \bf{58.94}  &                          \\ \cline{2-9}
& HYDRA~\cite{sehwag2020hydraPruning}         & \bf{88.98}& 60.13   & 57.66  & 58.42 & 65.01  & 57.64     & \multirow{2}{*}{+2.74}   \\
& HYDRA + TTE                                 & 88.82	    & 62.82   &	60.40  & 60.91 & 66.03  & \bf{60.38}&                          \\ \cline{2-9}
& AWP~\cite{wu2020adversarial}                & \bf{88.25}& 63.81   & 60.53  & 60.98 & 66.18  & 60.53     & \multirow{2}{*}{+1.46}   \\
& AWP + TTE                                   & 88.07	    & 64.95   &	61.99  & 62.52 & 66.48  & \bf{61.99}&                          \\ \cline{2-9}
& Gowal \etal\cite{gowal2020uncovering}       & \textbf{89.48}       & 66.16   & 63.26  & 63.74 & 69.10 & 63.29 & \multirow{2}{*}{+1.26} \\
& Gowal \etal + TTE                           & 89.41       & 67.19   & 64.55  & 64.88 & 69.29 & \textbf{64.55} & \\ \cline{2-9}

\hline\hline
\parbox[t]{2mm}{\multirow{6}{*}{\rotatebox[origin=c]{90}{CIFAR100}}} & ATES~\cite{sitawarin2020improving}          & 62.82	    & 26.78   &	24.98  & 25.23 & 31.27  & 24.96     & \multirow{2}{*}{+1.83} & \multirow{6}{*}{+1.29}  \\
& ATES + TTE                                  & \bf{63.47}& 28.9    &	26.8   & 27.15 & 32.21  & \bf{26.79}&                          \\ \cline{2-9}
& IN-Pret~\cite{hendrycks2019using}   & 59.37     & 33.45   & 29.03  & 29.34 & 34.55  & 28.61     & \multirow{2}{*}{+1.19}   \\
& IN-Pret + TTE                       & \bf{59.38}& 33.96   & 29.59  & 29.87 & 34.86  & \bf{29.50}&                          \\ \cline{2-9}
& AWP~\cite{wu2020adversarial}                & 60.38     & 33.56   & 29.16  & 29.48 & 34.66  & 29.15     & \multirow{2}{*}{+0.86}   \\
& AWP + TTE                                   & \bf{60.39}& 34.11   &	30.03  & 30.26 & 34.64	& \bf{30.01}&                          \\ \cline{1-10}
\end{tabular}
\label{tab:cifar10}
\end{table*}

%% file: tables/ImageNet.tex
\begin{table*}[]
\centering
\caption{\textbf{Adversarial robustness on ImageNet.} We compare the standard and the best TTE-enhanced versions of Feature Denoising~\cite{Xie2019CVPR} on the ImageNet dataset. We report the clean accuracy, the adversarial accuracy against individual attacks comprising AutoAttack, and the worst-case (\textit{Robust}) accuracy in $(\%)$. Following the experimental setup in~\cite{Xie2019CVPR}, we run this experiment with $\epsilon = \nicefrac{16}{255}$.}
\centering
\small
\begin{tabular}{l|c|cccc|c|c}
\hline
Method                  & Clean         & APGD-CE   & APGD-T    & FAB-T     & Square    & Robust    & Difference               \\ \hline\hline
FD~\cite{Xie2019CVPR}   & 65.32         & 7.91      & 4.31      & 7.87      & 23.76     & 4.23      & \multirow{2}{*}{+2.21}   \\
FD + TTE                & \bf{65.87}    & 9.29      & 6.53      & 9.23      & 26.34     & \bf{6.44} &                          \\ \hline
\end{tabular}
\label{tab:imagenet}
\end{table*}

%% file: tables/ablation_on_several_transformations.tex
\begin{table}[]
\centering
\caption{\textbf{Adversarial robustness gains of various transforms on CIFAR10.} We test the impact in adversarial robustness of introducing various transforms to TTE on TRADES. We report clean and robust accuracies, and the difference in robustness between each TTE-enhanced model and TRADES. Robust accuracies larger than that of TRADES are shown in \textbf{boldface}.}
\centering
\small
\begin{tabular}{l|c|c|c}
\hline
Method                                  & Clean      & Robust     & Diff.\\ \hline\hline
TRADES                                  & 84.92	     & 53.11      & -         \\ \hline
+ flip                                  & 85.07      & \bf{54.81} & +1.70     \\ \hline
+ 1 crop                                & 84.86      & \bf{53.35} & +0.24     \\ 
+ 2 crops                               & 84.85      & \bf{53.35} & +0.24     \\ 
+ 3 crops                               & 84.89      & \bf{53.63} & +0.52     \\ 
+ 4 crops                               & 84.87      & \bf{53.34} & +0.23     \\ \hline
+ flip + 1 crop                         & 85.09      & \bf{55.08} & +1.97     \\ 
+ flip + 2 crops                        & 84.82      & \bf{54.68} & +1.57     \\ 
+ flip + 3 crops                        & 85.07      & \bf{54.76} & +1.65     \\ 
+ flip + 4 crops                        & 85.11      & \bf{54.43} & +1.32     \\ \hline
+ flip + 1 crop + 1 flipped-crop        & 85.13      & \bf{55.19} & +2.08     \\ 
+ flip + 2 crops + 2 flipped-crops      & 85.06      & \bf{55.24} & +2.13     \\ 
+ flip + 3 crops + 3 flipped-crops      & 85.14      & \bf{55.49} & +2.38     \\ 
+ flip + 4 crops + 4 flipped-crops      & 85.13      & \bf{55.49} & +2.38     \\ \hline

\end{tabular}
\label{tab:transformations}
\end{table}

%% file: tables/ImageNet-transforms.tex
\begin{table}[]
\centering
\caption{\textbf{Adversarial robustness gains of various transforms on ImageNet.} We test the impact in adversarial robustness of introducing various transforms to TTE on Feature Denoising (FD)~\cite{Xie2019CVPR}. We report clean and PGD$^{30}$ accuracies, and the difference in robustness between each TTE-enhanced model and FD. PGD$^{30}$ accuracies larger than that of FD are shown in \textbf{boldface}.}
\centering
\small
\begin{tabular}{l|c|c|c}
\hline
Method                                  & Clean      & PGD$^{30}$      & Diff.\\ \hline\hline
FD~\cite{Xie2019CVPR}                   & 65.32	     & 50.20	   &  -     \\ \hline
+ flip                                  & 65.38      & \bf{51.17} & +0.97  \\ \hline
+ 1 crop                                & 65.50      & \bf{51.07} & +0.87  \\ 
+ 2 crops                               & 65.51      & \bf{50.84} & +0.64  \\ 
+ 3 crops                               & 65.78      & \bf{51.20} & +1.00  \\ 
+ 4 crops                               & 65.74      & \bf{51.21} & +1.01  \\ \hline
+ flip + 1 crop                         & 65.56      & \bf{51.69} & +1.49  \\ 
+ flip + 2 crops                        & 65.59      & \bf{51.77} & +1.57  \\ 
+ flip + 3 crops                        & 65.81      & \bf{51.80} & +1.60  \\ 
+ flip + 4 crops                        & 65.76      & \bf{51.43} & +1.23  \\ \hline
+ flip + 1 crop + 1 flipped-crop        & 65.69      & \bf{51.47} & +1.27  \\ 
+ flip + 2 crops + 2 flipped-crops      & 65.68      & \bf{51.36} & +1.15  \\ 
+ flip + 3 crops + 3 flipped-crops      & 65.87      & \bf{51.88} & +1.68  \\ 
+ flip + 4 crops + 4 flipped-crops      & 65.85      & \bf{52.17} & +1.98  \\ \hline

\end{tabular}
\label{tab:imagenet-transformations}
\end{table}

%% file: tables/obfuscation_ablations.tex
\begin{table}[]
\centering
\caption{\textbf{Accuracy under \textit{APGD-T} attacks \textit{vs.} optimization iterations and attack strength.} We study how accuracy under \textit{APGD-T} attacks changes as we vary \textit{(i)} iterations for optimization (top table), and \textit{(ii)} attack strength (bottom table). Our results suggest that TTE does \textit{not} induce gradient obfuscation.}
\centering
\small
\begin{tabular}{l|c|c|c|c}
\multicolumn{5}{c}{\textbf{Optimization iterations}} \\\hline
Iterations      & 5         & 10        & 50        & 100       \\ \hline\hline
TRADES          & 49.92     & 49.12     & 48.71     & 48.69     \\ 
TRADES + TTE    & \bf{52.11}& \bf{51.54}& \bf{51.41}& \bf{51.40}     \\  \hline
\end{tabular}

\begin{tabular}{l|c|c|c|c}
\multicolumn{5}{c}{\textbf{Attack strength ($\epsilon$)}} \\\hline
$\epsilon$      & $\nicefrac{8}{255}$   & $\nicefrac{16}{255}$      & $\nicefrac{32}{255}$  & $\nicefrac{64}{255}$  \\ \hline\hline
TRADES          & 48.69                 & 15.84                     & 0.72                  & 0.00                  \\ 
TRADES + TTE    & \bf{51.40}            & \bf{18.85}                & \bf{0.95}             & \bf{0.01}                  \\  \hline
\end{tabular}
\label{tab:grad_obf}
\end{table}

%% file: tables/without_pad_and_crop.tex
\begin{table}[]
\centering
\caption{\textbf{Matching train- and test-time transforms.} We introduce a mismatch between train- and test-time transforms and record variations in performance. We induce a mismatch by \textit{(i)} testing with TTE including Gaussian-filtering (top sub-table), and \textit{(ii)} training a TRADES model without the pad-and-crop transformation (TRADES$^\text{nc}$) and testing with TTE including pad-and-crop transforms (bottom sub-table). Results on CIFAR10 show that TTE's boosts require matching train- and test-time transforms.}
\centering
\small
\begin{tabular}{l|c|c|c}
\hline
Method                          & Clean     & Robust & Diff.    \\ \hline
TRADES                          & 84.92	    & 53.11  & -        \\ 
\hline
\hline
+ Gaussian ($k=3$, $\sigma=1$)  & 81.59     & 50.21  & -2.90    \\
+ Gaussian ($k=3$, $\sigma=2$)  & 81.27     & 49.63  & -3.48    \\ \hline
+ Gaussian ($k=5$, $\sigma=1$)  & 80.36     & 49.19  & -3.92    \\
+ Gaussian ($k=5$, $\sigma=2$)  & 76.78     & 45.71  & -7.40    \\ 
\hline\hline
TRADES$^\text{nc}$              & 82.40	    & 48.64  & -4.47    \\ 
TRADES$^\text{nc}$ + 1 crop     & 81.28	    & 47.04  & -6.07    \\ 
TRADES$^\text{nc}$ + 2 crops    & 77.03	    & 42.23  & -10.88   \\ 
TRADES$^\text{nc}$ + 3 crops    & 80.66	    & 46.57  & -6.54    \\ 
TRADES$^\text{nc}$ + 4 crops    & 77.79	    & 41.39  & -11.72   \\ \hline
\end{tabular}
\label{tab:nocrop_no_gauss}
\end{table}

%% file: Sections/conclusions.tex
\section{Conclusions}
In this work, we analyzed the effect of using Test-time Transformation Ensembling (TTE) on adversarial robustness. 
We conducted a comprehensive empirical study on the adversarial robustness effects of leveraging TTE through customary image transforms. Our results demonstrate that TTE is a simple yet effective technique for improving adversarial robustness. Notably, we showed that the performance of several SOTA adversarial robustness defenses can be boosted by including TTE, and, further, that these benefits transfer to the domain of certified defenses.

%% file: Sections/suppl.tex
\appendix
\section*{Supplementary Material}
Here we present the \textbf{Supplementary Material} for the paper \textit{Enhancing Adversarial Robustness via Test-time Transformation Ensembling}.
In this document, we report comprehensive results for ablations, qualitative results, pseudo-code for the transforms we used, plots for the complete outcomes of gradient-obfuscation experiments, and results for the removal of another common transform from training.

\section{Detailed Transforms Ablation}
In Tables~\ref{tab:cifar10} and \ref{tab:imagenet} we reported our best results for the transforms with which TTE is instantiated. However, Table~\ref{tab:transformations} only reports results in the case of TRADES on CIFAR10. Here, for completeness, we show the analogous results for the rest of the methods reported in Table~\ref{tab:cifar10} and for a large set of transforms.

We report CIFAR10-only methods (HYDRA, MART and Gowal \etal) in Tables~\ref{app:tab:hydra}, \ref{app:tab:MART} and \ref{app:tab:gowal}. We report methods with results both on CIFAR10 and CIFAR100 (AWP, ATES, and IN-Pret) in Tables \ref{app:tab:AWP:cif10}, \ref{app:tab:AWP:cif100}, \ref{app:tab:ATES:cif10}, \ref{app:tab:ATES:cif100}, \ref{app:tab:inpret:cif10}, and \ref{app:tab:inpret:cif100}.

\input{tables/appendix_GOWAL}

\input{tables/appendix_HYDRA}
\input{tables/appendix_MART}
\input{tables/appendix_AWP}
\input{tables/appendix_ATES}
\input{tables/appendix_INPret}

\section{Detailed Algorithms for Transforms}
Algorithm \ref{alg:transforms} reports the pseudo-code, detailing our implementation of the transforms used in our TTE wrapper.
\input{algorithms/transforms}

\section{Varying attack strength and iterations}
\begin{figure}[ht]
    \centering
    \includegraphics[width=1.0 \columnwidth]{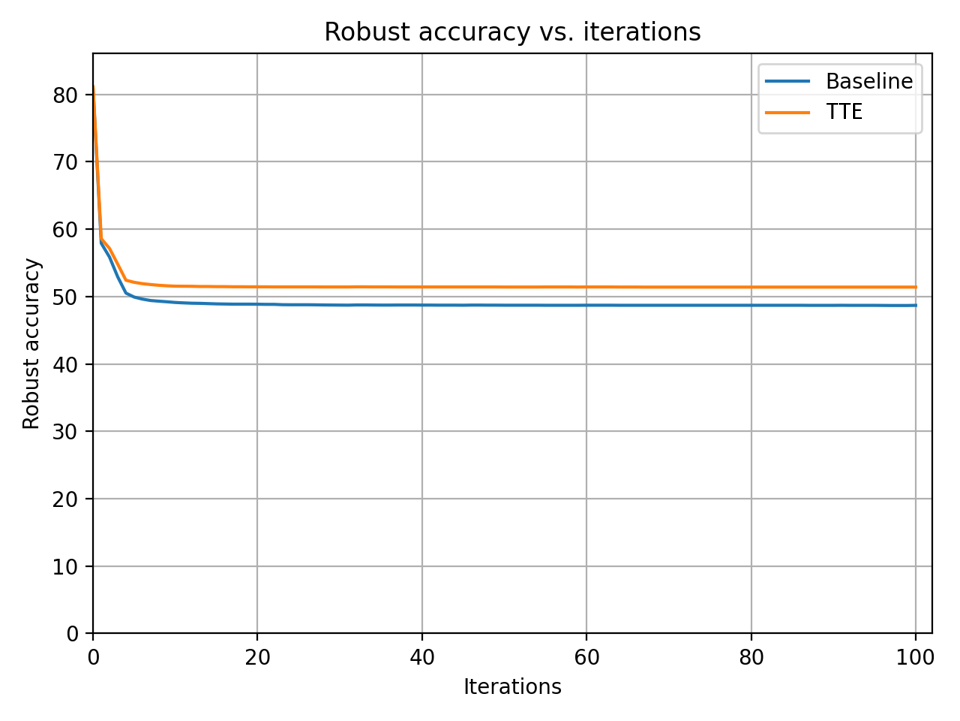}
    \includegraphics[width=1.0 \columnwidth]{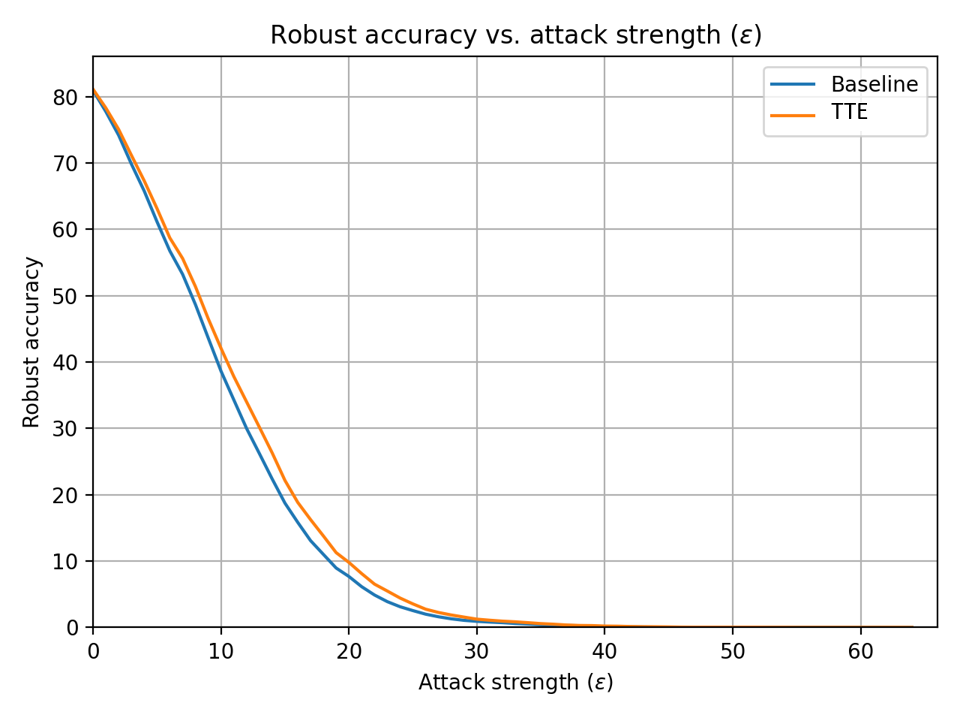}
    \caption{\textbf{Accuracy under APGD-T attacks \textit{vs.} optimizations iterations and attack strength.} We report accuracy plots for the entire set of values considerd in the experiments from Section~\ref{sec:obf}.}
    \label{fig:strength_iters}
\end{figure}

In Section~\ref{sec:obf} we conducted experiments on \textit{APGD-T} by varying the number of optimization iterations and the attack strength ($\epsilon$) to check for signs of gradient obfuscation. We reported accuracy under attack at important milestones of each of these ablations. Here we report accuracies for \textit{all} possible values in Figure~\ref{fig:strength_iters}. These results are consistent with the hypothesis that introducing TTE does not induce gradient obfuscation, as stated in the paper.

\section{Removing flip transform from training} 
Following a similar spirit to Section~\ref{sec:match}, we \textit{remove} the usual flipping transform from the official TRADES training routine, and refer to this model as TRADES$^\text{nf}$. We record the adversarial robustness of TRADES$^\text{nf}$ both when tested on \textit{(i)} clean images and \textit{(ii)} on both the original image and its flipped version. Table~\ref{tab:noflip} reports our results. From this table we note that \textit{(i)} the training-time flip transform is essential for TRADES: clean and robust accuracies drop, approximately, by $3$\% and $7$\%, respectively; \textit{(ii)} even with TRADES$^\text{nf}$, adding a flipped version of the image is beneficial for adversarial robustness: clean and robust accuracies increase, approximately, by $1$\% and $4$\%, respectively. These results suggest that, for a TRADES model, there is little distribution shift between the original images and their flipped versions, as expected. Thus, adding a flipped version of the image at test-time does not induce vulnerabilities into the defense.

\input{tables/without_flipping}

\section{Visualization}
To exemplify what the model receives as input when equipped with the TTE wrapper, we illustrate the four most complete cases from our ablations in Section A. That is, when the model is fed with flipped, cropped and flipped-cropped versions of the image, in addition to the original image. These cases are depicted in Figure X.

\begin{figure}[ht]
    \centering
    \includegraphics[width=1.0 \columnwidth]{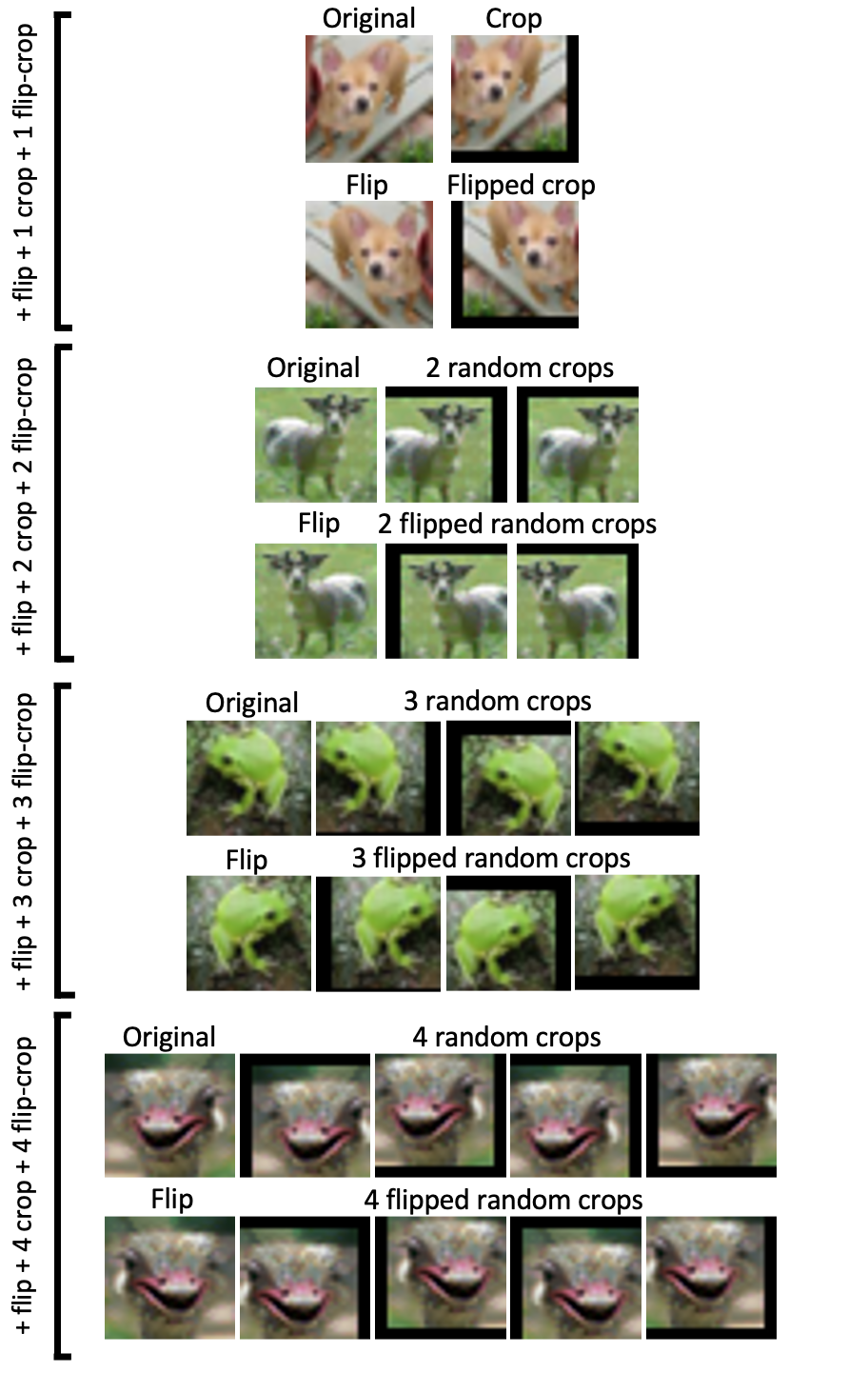}
    \caption{\textbf{Visualization of what an TTE-enhanced model receives as input.} We show what an input looks like to a model that has been enhanced with TTE. Here we exemplify what several sets of transforms result in.}
    \label{fig:crops}
    \vspace{-1.1em}
\end{figure}

\begin{figure}
    \centering
    \includegraphics[width=0.9\columnwidth]{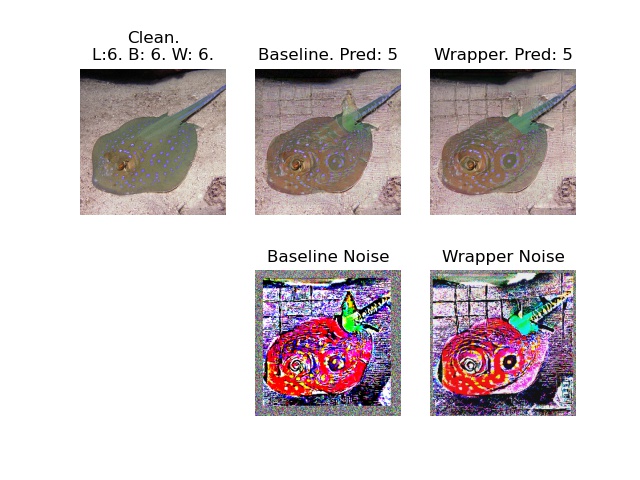}
    \caption{Original Image is labeled as \textit{Stingray}. The adversaries are predicted as \textit{Electric Rays}. Some electric rays are characteristic of having circular patterns like the ones induced by the adversarial noise.}
    \label{fig:adv0-300}
    \vspace{-0.2cm}
\end{figure}

\begin{figure}
    \centering
    \includegraphics[width=0.9\columnwidth]{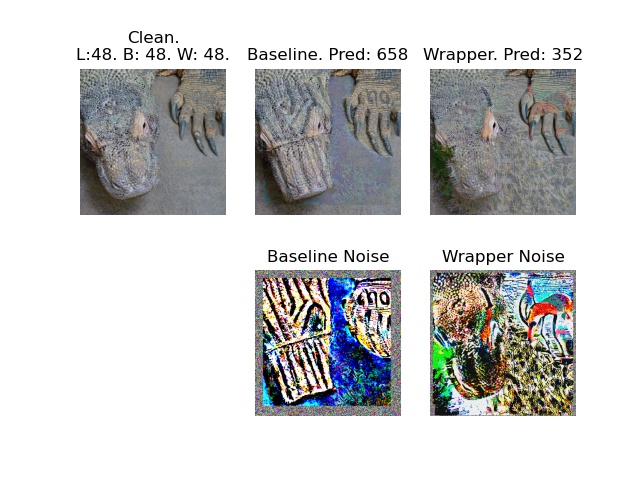}
    \caption{Original Image is labeled as \textit{Komodo Dragon}. The baseline model classifies its adversary as a \textit{Mitten}. The adversary of the wrapper, classified as \textit{Impala}, needed to modify all the image to fool the network.}
    \label{fig:adv4-0}
    \vspace{-0.2cm}
\end{figure}

\begin{figure}
    \centering
    \includegraphics[width=0.9\columnwidth]{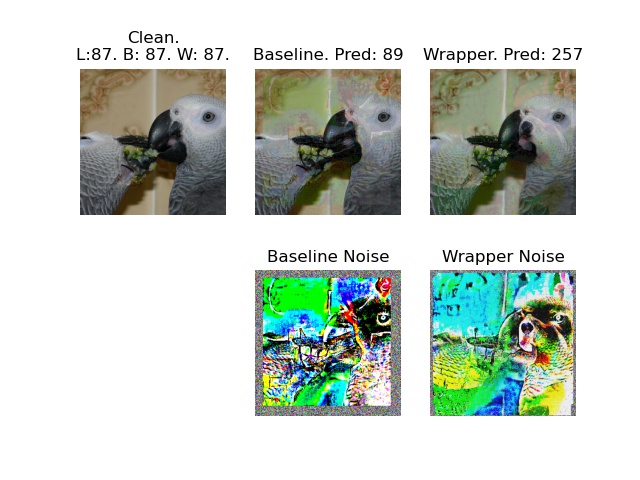}
    \caption{Original Image is labeled as \textit{African Grey Parrot}. The baseline model classifies its adversary as \textit{Sulphur-Crested Cockatoo}. The wrapper-enhanced version classifies its adversary as a \textit{Great Pyrenees}. The noise clearly displays the nose of this dog breed.}
    \label{fig:adv7-150}
    \vspace{-0.2cm}
\end{figure}

\begin{figure}
    \centering
    \includegraphics[width=0.9\columnwidth]{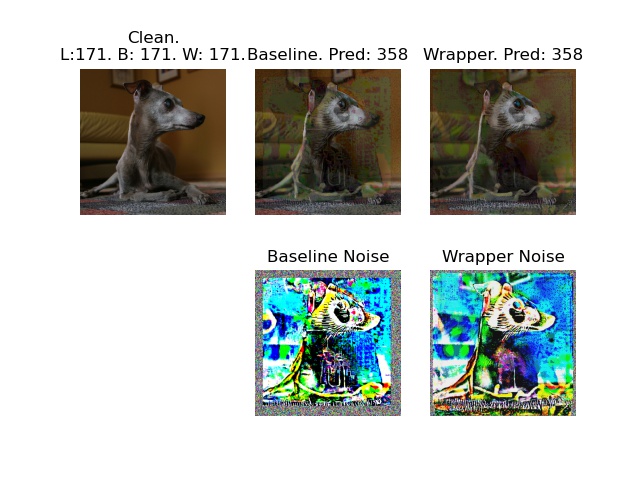}
    \caption{Original Image is labeled as \textit{Italian Greyhound}. Both models classify their adversarial examples as \textit{Polecat}. The noise from the wrapper-enhanced version clearly displays less blurry patterns.}
    \label{fig:adv14-150}
    \vspace{-0.2cm}
\end{figure}

\begin{figure}
    \centering
    \includegraphics[width=0.9\columnwidth]{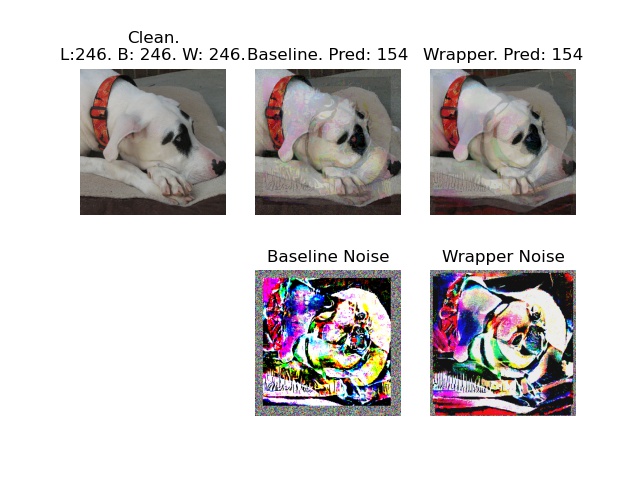}
    \caption{Original Image is labeled as \textit{Great Dane}. Both models classify their adversarial examples as \textit{Pekinese}. The noise from the wrapper-enhanced version clearly displays less blurry patterns.}
    \label{fig:adv20-300}
    \vspace{-0.2cm}
\end{figure}

\begin{figure}
    \centering
    \includegraphics[width=0.9\columnwidth]{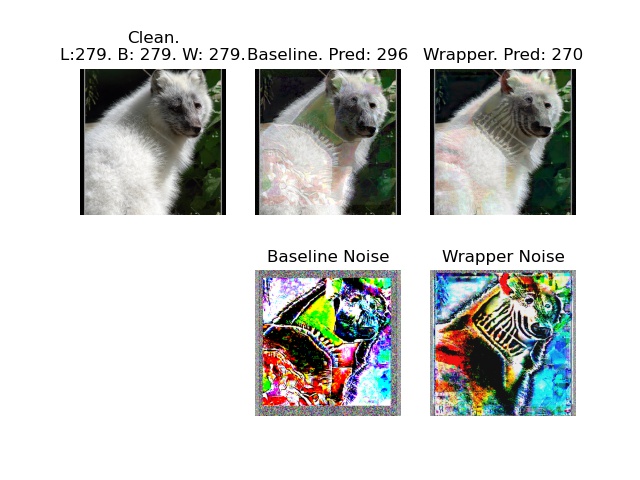}
    \caption{Original Image is labeled as \textit{Artic Fox}. The baseline model classify its adversarial example as \textit{Ice Bear} while the wrapper-enhanced version classifies his example as \textit{White Wolf}.}
    \label{fig:adv23-150}
    \vspace{-0.2cm}
\end{figure}

\begin{figure}
    \centering
    \includegraphics[width=0.9\columnwidth]{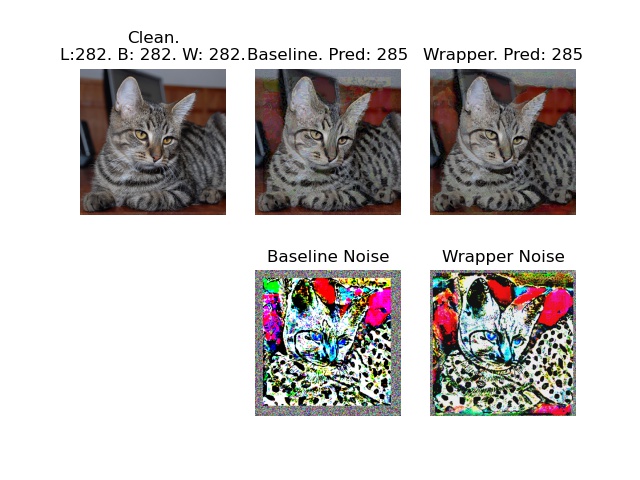}
    \caption{Original Image is labeled as \textit{Tiger Cat}. Both models classify their adversarial examples as \textit{Egyptian Cat}. The Egyptian cat is characterized by its dot marks.}
    \label{fig:adv23-300}
    \vspace{-0.2cm}
\end{figure}

\begin{figure}
    \centering
    \includegraphics[width=0.9\columnwidth]{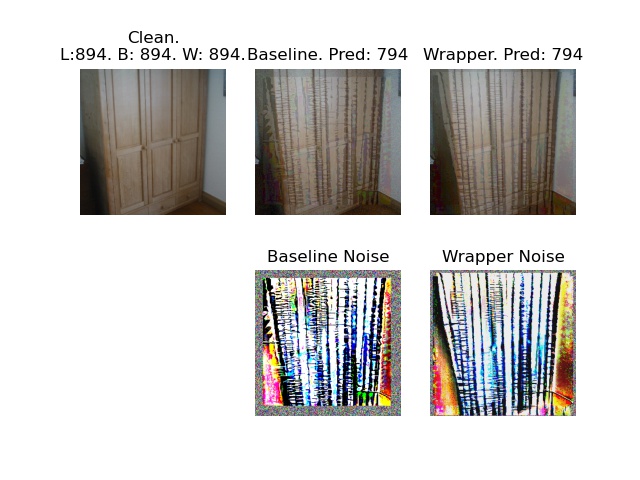}
    \caption{Original Image is labeled as \textit{Wardrobe}. Both models classify their adversarial examples as \textit{Shower Curtain}. The noise from the wrapper-enhanced version clearly displays less blurry patterns.}
    \label{fig:adv74-300}
    \vspace{-0.2cm}
\end{figure}

We also display some adversarial examples that fool the FD model (on ImageNet) with and without the wrapper enhancement. We display the adversarial examples and the noise introduced into the original image. For the baseline model, we extract the $224\times224$ center crop from the $256\times256$ crop. For this reason, there is a random-like padding as the initial perturbation is initialized from random noise. The adversarial examples are shown in Figures~\ref{fig:adv0-300} through~\ref{fig:adv74-300}.

\begin{figure}
    \centering
    \includegraphics[width=0.9\columnwidth]{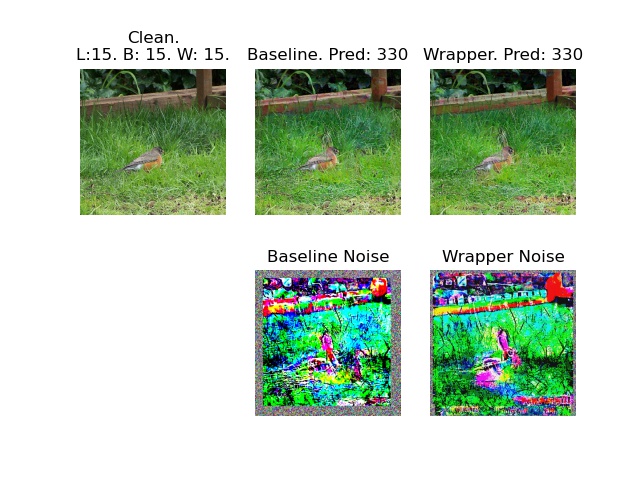}
    \caption{Original Image is labeled as \textit{Robin}. The adversaries are predicted as \textit{Wood Rabbits}. The adversarial example's noise from the wrapper-enhanced model clearly visualizes the ear from a bunny while the FD model does not exhibit this pattern.}
    \label{fig:adv1-150}
    \vspace{-0.2cm}
\end{figure}

\begin{figure}
    \centering
    \includegraphics[width=0.9\columnwidth]{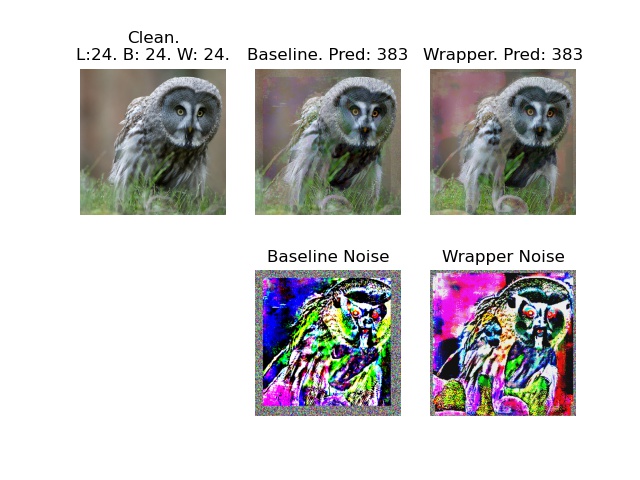}
    \caption{Original Image is labeled as \textit{Great Grey Owl}. The adversaries are predicted as \textit{Madagascar Cat}. The adversarial example's noise from the wrapper-enhanced model clearly visualizes the face from this animal while the FD model does not exhibit this pattern.}
    \label{fig:adv1-150}
    \vspace{-0.2cm}
\end{figure}

\begin{figure}
    \centering
    \includegraphics[width=0.9\columnwidth]{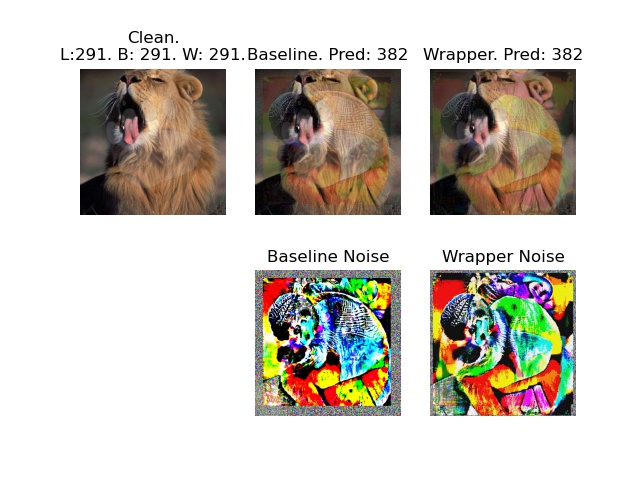}
    \caption{Original Image is labeled as \textit{Lion}. Both models classify their adversarial examples as \textit{Squirrel Monkey}. The shape of the monkey is clearly seen on the noise of both adversaries.}
    \label{fig:adv24-150}
    \vspace{-0.2cm}
\end{figure}

\begin{figure}
    \centering
    \includegraphics[width=0.9\columnwidth]{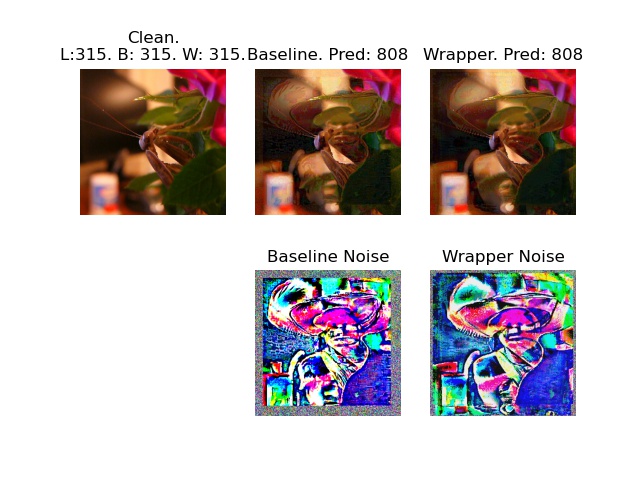}
    \caption{Original Image is labeled as \textit{Mantis}. Both models classify their adversarial examples as \textit{Sombrero}. Both adversaries display face-like attributes.}
    \label{fig:adv24-150}
    \vspace{-0.2cm}
\end{figure}

\begin{figure}
    \centering
    \includegraphics[width=0.9\columnwidth]{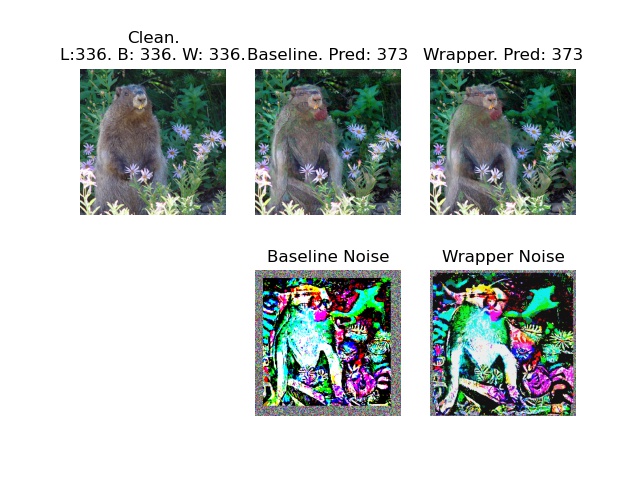}
    \caption{Original Image is labeled as \textit{Marmot}. Both models classify their adversarial examples as \textit{Macaque}. The noise from the wrapper-enhanced version clearly displays less blurry patterns.}
    \label{fig:adv24-150}
    \vspace{-0.2cm}
\end{figure}

\begin{figure}
    \centering
    \includegraphics[width=0.9\columnwidth]{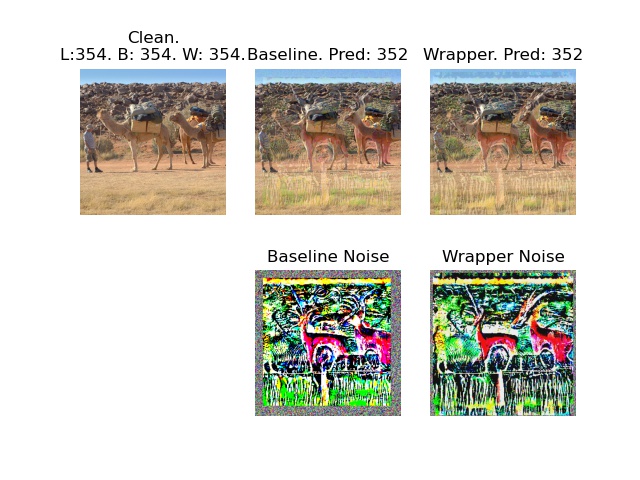}
    \caption{Original Image is labeled as \textit{Arabian Camel}. Both models classify their adversarial examples as \textit{Impala}. The wrapper-enhanced version exhibits clearer horns on the noise.}
    \label{fig:adv24-150}
    \vspace{-0.2cm}
\end{figure}

\begin{figure}
    \centering
    \includegraphics[width=0.9\columnwidth]{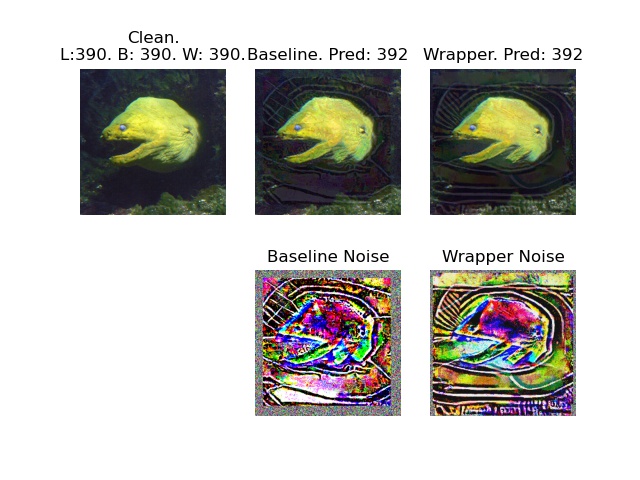}
    \caption{Original Image is labeled as \textit{Eel}. Both models classify their adversarial examples as \textit{Rock Beauty}. The wrapper-enhanced adversarial version exhibits clearer patterns.}
    \label{fig:adv24-150}
    \vspace{-0.2cm}
\end{figure}

\begin{figure}
    \centering
    \includegraphics[width=0.9\columnwidth]{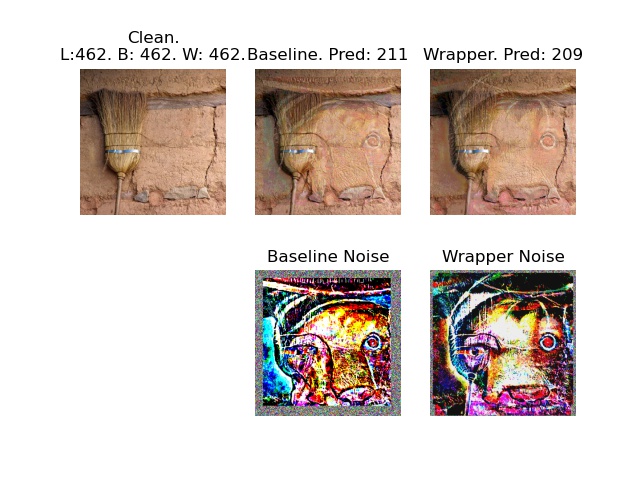}
    \caption{Original Image is labeled as \textit{Broom}. Both models classify their adversarial examples as dog breeds: \textit{Vizsla} for the baseline and \textit{Chesapeake Bay Retriever} for the wrapper version. The noise from both adversaries expose different shapes of the nose.}
    \label{fig:adv24-150}
    \vspace{-0.2cm}
\end{figure}

\begin{figure}
    \centering
    \includegraphics[width=0.9\columnwidth]{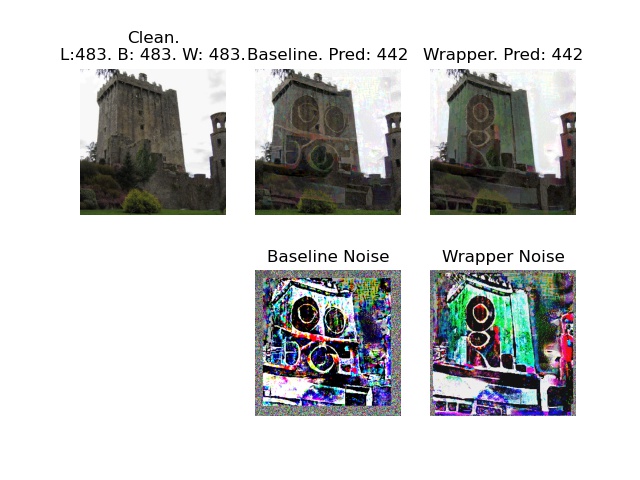}
    \caption{Original Image is labeled as \textit{Castle}. Both models classify their adversarial examples as \textit{Bell Cote}. The noise from the wrapper-enhanced version clearly displays less blurry patterns.}
    \label{fig:adv24-150}
    \vspace{-0.2cm}
\end{figure}

\begin{figure}
    \centering
    \includegraphics[width=0.9\columnwidth]{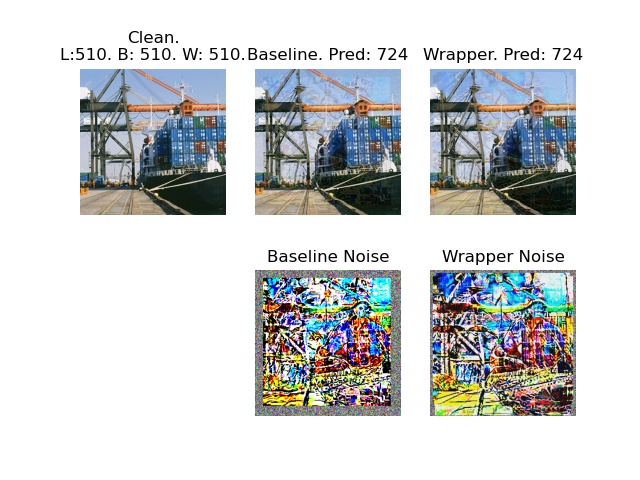}
    \caption{Original Image is labeled as \textit{Container Ship}. Both models classify their adversarial examples as \textit{Pirate}.}
    \label{fig:adv24-150}
    \vspace{-0.2cm}
\end{figure}

\begin{figure}
    \centering
    \includegraphics[width=0.9\columnwidth]{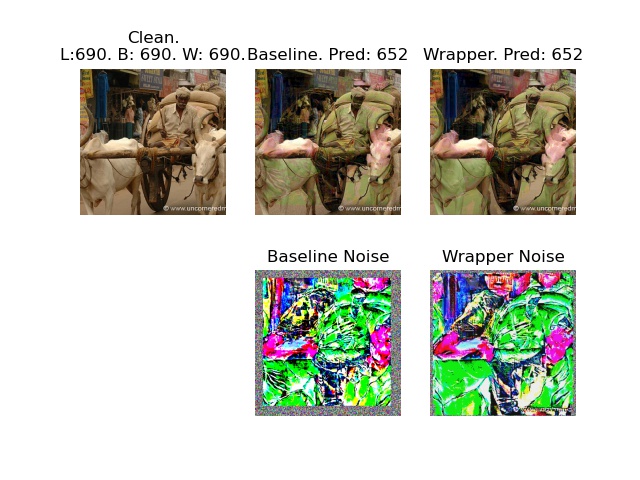}
    \caption{Original Image is labeled as \textit{Oxcart}. Both models classify their adversarial examples as \textit{Military Uniform}. The noise from the wrapper-enhanced version clearly displays less blurry patterns.}
    \label{fig:adv24-150}
    \vspace{-0.2cm}
\end{figure}

\begin{figure}
    \centering
    \includegraphics[width=0.9\columnwidth]{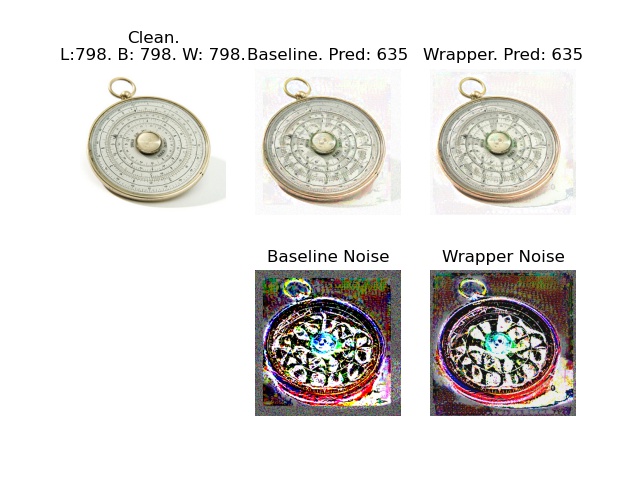}
    \caption{Original Image is labeled as \textit{Slide Rule}. Both models classify their adversarial examples as \textit{Magnetic Compass}. The noise from the wrapper-enhanced version clearly displays less blurry patterns.}
    \label{fig:adv24-150}
    \vspace{-0.2cm}
\end{figure}

%% file: tables/appendix_GOWAL.tex
\begin{table}[h]
    \centering
    \caption{\textbf{Adversarial robustness gains of various transforms on CIFAR10.} We test the impact in adversarial robustness of introducing various transforms to TTE on the method of Gowal \etal. We report clean and robust accuracies, and the difference in robustness between each TTE-enhanced model and Gowal \etal. Robust accuracies larger than that of Gowal \etal are shown in \textbf{boldface}.}
    \centering
    \small
    \begin{tabular}{l|c|c|c}
    \hline
    Method                                  & Clean      & Robust     & Diff.\\ \hline\hline
    Gowal \etal                             & 89.48	&	63.26      & -         \\ \hline
    + flip                                  & 89.41	&	\bf{64.37} & +1.09     \\ \hline
    + 1 crop                                & 89.39	&	\bf{63.52} & +0.26     \\ 
    + 2 crops                               & 89.04	&	     63.20 & -0.06     \\ 
    + 3 crops                               & 89.25 &   \bf{63.77} & +0.51     \\ 
    + 4 crops                               & 89.17	&	    63.22  & -0.04     \\ \hline
    + flip + 1 crop                         & 89.43	&	\bf{64.35} & +1.09     \\ 
    + flip + 2 crops                        & 89.16	&   \bf{64.12} & +0.86     \\ 
    + flip + 3 crops                        & 89.40	&	\bf{64.39} & +1.13    \\ 
    + flip + 4 crops                        & 89.18	&	\bf{63.95} & +0.69     \\ \hline
    + flip + 1 crop + 1 flipped-crop        & 89.49	&	\bf{64.40} & +1.14     \\ 
    + flip + 2 crops + 2 flipped-crops      & 89.05	&	\bf{64.20} & +0.94     \\ 
    + flip + 3 crops + 3 flipped-crops      & 89.41	&	\bf{64.55} & +1.29     \\ 
    + flip + 4 crops + 4 flipped-crops      & 89.21	 &  \bf{64.29} & +1.03     \\ \hline
    
    \end{tabular}
    \label{app:tab:gowal}
\end{table}

%% file: tables/appendix_HYDRA.tex
\begin{table}[h]
    \centering
    \caption{\textbf{Adversarial robustness gains of various transforms on CIFAR10.} We test the impact in adversarial robustness of introducing various transforms to TTE on HYDRA. We report clean and robust accuracies, and the difference in robustness between each TTE-enhanced model and HYDRA. Robust accuracies larger than that of HYDRA are shown in \textbf{boldface}.}
    \centering
    \small
    \begin{tabular}{l|c|c|c}
    \hline
    Method                                  & Clean      & Robust     & Diff.\\ \hline\hline
    HYDRA                                  & 88.98	&	57.64      & -         \\ \hline
    + flip                                  & 89.1	&	\bf{59.81}	& +2.17     \\ \hline
    + 1 crop                                & 88.86	&	\bf{58.36}	& +0.72     \\ 
    + 2 crops                               & 88.58	&	\bf{58.01}	& +0.37     \\ 
    + 3 crops                               & 88.92	&	\bf{58.61}	& +0.97     \\ 
    + 4 crops                               & 88.59	&	\bf{58.2}	& +0.56     \\ \hline
    + flip + 1 crop                         & 89.00   &	\bf{60.01}	& +2.37     \\ 
    + flip + 2 crops                        & 88.87	&	\bf{59.59}	& +1.95     \\ 
    + flip + 3 crops                        & 88.96	&	\bf{59.65}	& +2.01    \\ 
    + flip + 4 crops                        & 88.64	&	\bf{59.12}	& +1.48     \\ \hline
    + flip + 1 crop + 1 flipped-crop        & 88.89	&	\bf{60.28}	& +2.64     \\ 
    + flip + 2 crops + 2 flipped-crops      & 88.81	&	\bf{60.10}	& +2.46     \\ 
    + flip + 3 crops + 3 flipped-crops      & 88.82	&	\bf{60.38}	& +2.74     \\ 
    + flip + 4 crops + 4 flipped-crops      & 88.70	&	\bf{60.08}	& +2.44     \\ \hline
    
    \end{tabular}
    \label{app:tab:hydra}
\end{table}

%% file: tables/appendix_MART.tex
\begin{table}[]
    \centering
    \caption{\textbf{Adversarial robustness gains of various transforms on CIFAR10.} We test the impact in adversarial robustness of introducing various transforms to TTE on MART. We report clean and robust accuracies, and the difference in robustness between each TTE-enhanced model and MART. Robust accuracies larger than that of MART are shown in \textbf{boldface}.}
    \centering
    \small
    \begin{tabular}{l|c|c|c}
    \hline
    Method                                  & Clean      & Robust     & Diff.\\ \hline\hline
    MART                                  & 87.5	&56.75      & -         \\ \hline
    + flip                                  & 87.74	&	\bf{58.38}	&	+1.63     \\ \hline
    + 1 crop                                & 87.55	&	\bf{57.27}	&	+0.52     \\ 
    + 2 crops                               & 87.11	&	\bf{57.16}	&	+0.41     \\ 
    + 3 crops                               & 87.45	&	\bf{57.66}	&	+0.91     \\ 
    + 4 crops                               & 87.31	&	\bf{57.4}	&	+0.65     \\ \hline
    + flip + 1 crop                         & 87.66	&	\bf{58.53}	&	+1.78     \\ 
    + flip + 2 crops                        & 87.54	&	\bf{58.16}	&	+1.41     \\ 
    + flip + 3 crops                        & 87.69	&	\bf{58.42}	&	+1.67    \\ 
    + flip + 4 crops                        & 87.58	&	\bf{58.11}	&	+1.36     \\ \hline
    + flip + 1 crop + 1 flipped-crop        & 87.76	&	\bf{58.83}	&	+2.08     \\ 
    + flip + 2 crops + 2 flipped-crops      & 87.61	&	\bf{58.87}	&	+2.12     \\ 
    + flip + 3 crops + 3 flipped-crops      & 87.79	&	\bf{58.94}	&	+2.19     \\ 
    + flip + 4 crops + 4 flipped-crops      & 87.61	&	\bf{58.92}	&	+2.17     \\ \hline
    
    \end{tabular}
    \label{app:tab:MART}
\end{table}

%% file: tables/appendix_AWP.tex
\begin{table}[h]
    \centering
    \caption{\textbf{Adversarial robustness gains of various transforms on CIFAR10.} We test the impact in adversarial robustness of introducing various transforms to TTE on AWP. We report clean and robust accuracies, and the difference in robustness between each TTE-enhanced model and AWP. Robust accuracies larger than that of AWP are shown in \textbf{boldface}.}
    \centering
    \small
    \begin{tabular}{l|c|c|c}
    \hline
    Method                                  & Clean      & Robust     & Diff.\\ \hline\hline
    AWP                                  & 88.25	&	60.53      & -         \\ \hline
    + flip                                  & 88.2	&	\bf{61.54} & +1.01     \\ \hline
    + 1 crop                                & 88.08	&	\bf{60.82} & +0.29     \\ 
    + 2 crops                               & 87.76	&	60.42 & -0.11     \\ 
    + 3 crops                               & 88.04 &   \bf{60.99} & +0.46     \\ 
    + 4 crops                               & 87.87	&	\bf{60.79} & +0.26     \\ \hline
    + flip + 1 crop                         & 88.28	&	\bf{61.71} & +1.18     \\ 
    + flip + 2 crops                        & 87.92	&   \bf{61.35} & +0.82     \\ 
    + flip + 3 crops                        & 88.08	&	\bf{61.6} & +1.07    \\ 
    + flip + 4 crops                        & 88.03	&	\bf{61.29} & +0.76     \\ \hline
    + flip + 1 crop + 1 flipped-crop        & 88.23	&	\bf{61.68} & +1.15     \\ 
    + flip + 2 crops + 2 flipped-crops      & 88.06	&	\bf{61.54} & +1.01     \\ 
    + flip + 3 crops + 3 flipped-crops      & 88.07	&	\bf{61.99} & +1.46     \\ 
    + flip + 4 crops + 4 flipped-crops      & 87.98	 &   \bf{61.62} & +1.09     \\ \hline
    
    \end{tabular}
    \label{app:tab:AWP:cif10}
\end{table}

\begin{table}[h]
    \centering
    \caption{\textbf{Adversarial robustness gains of various transforms on CIFAR100.} We test the impact in adversarial robustness of introducing various transforms to TTE on AWP. We report clean and robust accuracies, and the difference in robustness between each TTE-enhanced model and AWP. Robust accuracies larger than that of AWP are shown in \textbf{boldface}.}
    \centering
    \small
    \begin{tabular}{l|c|c|c}
    \hline
    Method                                  & Clean      & Robust     & Diff.\\ \hline\hline
    AWP		                  &  60.38	&28.86	& - \\ \hline
    +flip	    	                  &  60.27	&\bf{29.66}	&+0.80\\ \hline
    +1 crop  		                  &  60.48	&\bf{29.31}	&+0.45\\
    +2 crops	                         &	59.96	&\bf{29.29}	&+0.43\\
    +3 crops	                        &	60.41	&\bf{29.53}	&+0.67\\
    +4 crops	                       & 	60.3	&\bf{29.41}	&+0.55\\ \hline
    +flip + 1 crop		           &     60.36	&\bf{29.79}	&+0.93\\
    +flip + 2 crop               &		60.26	&\bf{29.70}	&+0.84\\
    +flip + 3 crop		              &  60.43	&\bf{29.80} &+0.94\\
    +flip + 4 crop		             &   60.44	&\bf{29.74}	&+0.88\\ \hline
    +flip + 1 crop + 1 flipped-crop	&	60.28	&\bf{29.86}	&+1.00\\
    +flip + 2 crop + 2 flipped-crop	&	60.22	&\bf{29.94}	&+1.08\\
    +flip + 3 crop + 3 flipped-crop	&	60.39	&\bf{30.01}	&+1.15\\
    +flip + 4 crop + 4 flipped-crop		&60.13	&\bf{29.78}	&+0.92\\ \hline
    
    \end{tabular}
    \label{app:tab:AWP:cif100}
\end{table}

%% file: tables/appendix_ATES.tex
\begin{table}[h]
    \centering
    \caption{\textbf{Adversarial robustness gains of various transforms on CIFAR10.} We test the impact in adversarial robustness of introducing various transforms to TTE on ATES. We report clean and robust accuracies, and the difference in robustness between each TTE-enhanced model and ATES. Robust accuracies larger than that of ATES are shown in \textbf{boldface}.}
    \centering
    \small
    \begin{tabular}{l|c|c|c}
    \hline
    Method                                  & Clean      & Robust     & Diff.\\ \hline\hline
    ATES                                    & 86.84	&	51.46      & -         \\ \hline
    + flip                                  & 86.96	&	\bf{53.11} & +1.65     \\ \hline
    + 1 crop                                & 86.86	&	\bf{52.08} & +0.62     \\ 
    + 2 crops                               & 86.68	&	\bf{52.37} & +0.91     \\ 
    + 3 crops                               & 86.86 &   \bf{52.59} & +1.13     \\ 
    + 4 crops                               & 86.62	&	\bf{52.31} & +0.85     \\ \hline
    + flip + 1 crop                         & 86.96	&	\bf{53.26} & +1.80     \\ 
    + flip + 2 crops                        & 86.89	&   \bf{53.54} & +2.08     \\ 
    + flip + 3 crops                        & 87.03	&	\bf{53.46} & +2.00    \\ 
    + flip + 4 crops                        & 86.82	&	\bf{53.25} & +1.79     \\ \hline
    + flip + 1 crop + 1 flipped-crop        & 87.08	&	\bf{53.71} & +2.25     \\ 
    + flip + 2 crops + 2 flipped-crops      & 86.95	&	\bf{53.94} & +2.48     \\ 
    + flip + 3 crops + 3 flipped-crops      & 87.03	&	\bf{54.05} & +2.59     \\ 
    + flip + 4 crops + 4 flipped-crops      & 86.86	 &   \bf{54.17} & +2.71     \\ \hline
    
    \end{tabular}
    \label{app:tab:ATES:cif10}
\end{table}

\begin{table}[h]
    \centering
    \caption{\textbf{Adversarial robustness gains of various transforms on CIFAR100.} We test the impact in adversarial robustness of introducing various transforms to TTE on ATES. We report clean and robust accuracies, and the difference in robustness between each TTE-enhanced model and ATES. Robust accuracies larger than that of ATES are shown in \textbf{boldface}.}
    \centering
    \small
    \begin{tabular}{l|c|c|c}
    \hline
    Method                          & Clean     & Robust     & Diff.\\ \hline\hline
    ATES		                    &  62.82	&24.96	& - \\ \hline
    +flip	    	                &  63.11	&\bf{26.27}	& +1.31\\ \hline
    +1 crop  		                &  62.88	&\bf{25.77}	& +0.81\\
    +2 crops	                    &	62.70	&\bf{26.14}	& +1.18\\
    +3 crops	                    &	63.12	&\bf{26.07}	& +1.11\\
    +4 crops	                    & 	62.88	&\bf{25.75}	& +0.79\\ \hline
    +flip + 1 crop		            &   63.27	&\bf{26.45}	& +1.49\\
    +flip + 2 crop                  &	62.70	&\bf{26.14}	& +1.18\\
    +flip + 3 crop		            &  63.20	&\bf{26.61} & +1.65\\
    +flip + 4 crop		            &   63.21	&\bf{26.43}	& +1.47\\ \hline
    +flip + 1 crop + 1 flipped-crop	&	63.17	&\bf{26.72}	& +1.76\\
    +flip + 2 crop + 2 flipped-crop	&	62.97	&\bf{27.04}	& +2.08\\
    +flip + 3 crop + 3 flipped-crop	&	63.47	&\bf{26.79}	& +1.83\\
    +flip + 4 crop + 4 flipped-crop	&   63.24	&\bf{27.09}	& +2.13\\ \hline
    
    \end{tabular}
    \label{app:tab:ATES:cif100}
\end{table}

%% file: tables/appendix_INPret.tex
\begin{table}[h]
    \centering
    \caption{\textbf{Adversarial robustness gains of various transforms on CIFAR10.} We test the impact in adversarial robustness of introducing various transforms to TTE on IN-Pret. We report clean and robust accuracies, and the difference in robustness between each TTE-enhanced model and IN-Pret. Robust accuracies larger than that of IN-Pret are shown in \textbf{boldface}.}
    \centering
    \small
    \begin{tabular}{l|c|c|c}
    \hline
    Method                                  & Clean     & Robust     & Diff.\\ \hline\hline
    IN-Pret                                 & 87.11	    &	55.31      & -         \\ \hline
    + flip                                  & 87.06	    &	\bf{55.66} & +0.35     \\ \hline
    + 1 crop                                & 87.23	    &	\bf{56.20} & +0.89     \\ 
    + 2 crops                               & 86.74     &	\bf{55.34} & +0.03     \\ 
    + 3 crops                               & 86.96     &   \bf{55.67} & +0.36     \\ 
    + 4 crops                               & 86.71     &	55.06 & -0.25     \\ 
    \hline
    + flip + 1 crop                         & 87.22     &	\bf{56.45}  & +1.14     \\ 
    + flip + 2 crops                        & 86.85	    &   \bf{56.11}  & +0.80     \\ 
    + flip + 3 crops                        & 87.06	    &	\bf{56.24}  & +0.93    \\ 
    + flip + 4 crops                        & 86.84     &	\bf{55.87} & +0.56     \\ \hline
    + flip + 1 crop + 1 flipped-crop        & 87.13	    &	\bf{56.43} & +1.12     \\ 
    + flip + 2 crops + 2 flipped-crops      & 86.93	    &	\bf{56.49} & +1.18     \\ 
    + flip + 3 crops + 3 flipped-crops      & 87.17	    &	\bf{56.50} & +1.19     \\ 
    + flip + 4 crops + 4 flipped-crops      & 86.85	    &   \bf{56.41} & +1.10     \\ \hline
    
    \end{tabular}
    \label{app:tab:inpret:cif10}
\end{table}

\begin{table}[h]
    \centering
    \caption{\textbf{Adversarial robustness gains of various transforms on CIFAR100.} We test the impact in adversarial robustness of introducing various transforms to TTE on IN-Pret. We report clean and robust accuracies, and the difference in robustness between each TTE-enhanced model and IN-Pret. Robust accuracies larger than that of IN-Pret are shown in \textbf{boldface}.}
    \centering
    \small
    \begin{tabular}{l|c|c|c}
    \hline
    Method                          & Clean     & Robust     & Diff.\\ \hline\hline
    IN-Pret		                    &  59.37	&28.96	& - \\ \hline
    +flip	    	                &  59.52	&\bf{29.40}	& +0.44\\ \hline
    +1 crop  		                &  58.96	&\bf{29.02}	& +0.06\\
    +2 crops	                    &  58.64	&28.67	& -0.29\\
    +3 crops	                    &  59.15	&\bf{29.10} & +0.14\\
    +4 crops	                    &  58.61    &28.67	& -0.29\\
    \hline
    +flip + 1 crop		            &   59.39	&\bf{29.46}	& +0.50\\
    +flip + 2 crop                  &   58.94   &\bf{29.24}	& +0.28\\
    +flip + 3 crop		            &   59.10   &\bf{29.39} & +0.43\\
    +flip + 4 crop		            &   58.87   &\bf{29.12} & +0.16\\ \hline
    +flip + 1 crop + 1 flipped-crop	&	59.38	&\bf{29.50}	& +0.54\\
    +flip + 2 crop + 2 flipped-crop	&	58.75	&\bf{29.32}	& +0.36\\
    +flip + 3 crop + 3 flipped-crop	&	59.16	&\bf{29.61}	& +0.65\\
    +flip + 4 crop + 4 flipped-crop	&   58.93	&\bf{29.68}	& +0.72\\ \hline
    
    \end{tabular}
    \label{app:tab:inpret:cif100}
\end{table}

%% file: algorithms/transforms.tex
\begin{algorithm}[t]
\caption{Differentiable transforms pseudocode in PyTorch style.}
\label{alg:transforms}
\algcomment{\fontsize{7.2pt}{0em}\selectfont \texttt{pad}: zero padding.
}
\definecolor{codeblue}{rgb}{0.25,0.5,0.5}
\lstset{
  backgroundcolor=\color{white},
  basicstyle=\fontsize{7.2pt}{7.2pt}\ttfamily\selectfont,
  columns=fullflexible,
  breaklines=true,
  captionpos=b,
  commentstyle=\fontsize{7.2pt}{7.2pt}\color{codeblue},
  keywordstyle=\fontsize{7.2pt}{7.2pt},
}
\begin{lstlisting}[language=python]
class PadCrop:
    def __init__(self, o_x, o_y, crop_size, pad_size):
        self.pad_size = pad_size
        # starting points
        self.o_x = o_x
        self.o_y = o_y
        # ending points
        self.e_x = o_x + crop_size
        self.e_y = o_y + crop_size
    
    def forward(self, x):
        # pad input
        x = pad(x, pad=self.pad_size)
        # crop
        x = x[:,:,self.o_x:self.e_x,self.o_y:self.e_y]
        return x
        
class Flip:
    def forward(self, x):
        return x.flip(3) # the left-right dimension
        
class FlipPadCrop:
    def __init__(self, o_x, o_y, crop_size, pad_size):
        self.flip = Flip()
        self.pad_crop = PadCrop(o_x, o_y, crop_size, pad_size)
        
    def forward(self, x):
        return self.flip(self.pad_crop(x))
\end{lstlisting}
\end{algorithm}

%% file: tables/without_flipping.tex
\begin{table}[]
\centering
\caption{\textbf{Robustness of TRADES trained \textit{without} the flipping transformation.} We train a TRADES model without the flipping transformation (TRADES$^\text{nf}$). We test the model's adversarial robustness when tested on \textit{(i)} clean images and \textit{(ii)} on both the clean image and its flipped version. Results show that, even when the model was not trained on flipped images, introducing a flipped version of the image is still beneficial for adversarial robustness.}
\centering
\small
\begin{tabular}{l|l|l|c}
\hline
Method                      & Clean     & Robust & Diff.    \\ \hline\hline
TRADES                      & 84.92	    & 53.11  & -        \\
\hline
TRADES$^\text{nf}$          & 81.89	    & 46.29  & -6.82    \\ 
TRADES$^\text{nf}$ + flip   & 82.82	    & 50.19  & -2.92    \\ \hline
\end{tabular}
\label{tab:noflip}
\end{table}